\documentclass[lettersize,journal]{IEEEtran}
\usepackage{amsmath,amsfonts}
\usepackage{algorithmic}
\usepackage{algorithm}
\usepackage{array}
\usepackage[caption=false,font=normalsize,labelfont=sf,textfont=sf]{subfig}
\usepackage{textcomp}
\usepackage{stfloats}
\usepackage{url}
\usepackage{verbatim}
\usepackage{graphicx}
\usepackage{cite}
\usepackage{tikz, xcolor}
\usetikzlibrary{arrows.meta}
\definecolor{lime}{HTML}{A6CE39}
\definecolor{c1}{HTML}{d6ecf0}
\definecolor{metablue}{HTML}{0064E0}
\definecolor{metafg}{HTML}{1C2B33}
\definecolor{metabg}{HTML}{F1F4F7}

\usepackage{hyperref}
\hypersetup{
    colorlinks,
    citecolor=metablue,
}

\usepackage{booktabs}
\usepackage{tabularx}
\usepackage{multirow}
\usepackage{makecell}
\usepackage{colortbl}

\newcommand{\circlednum}[1]{%
    \tikz[baseline=(X.base)] {
        \node[draw,circle,inner sep=.3pt] (X) {#1};
    }%
}

\usepackage[capitalize]{cleveref}
\crefname{section}{Sec.}{Secs.}
\crefname{section}{Section}{Sections}
\crefname{table}{Tab.}{Tabs.}

\newcommand{\minisection}[1]{\vspace{0.04in} \noindent {\bf #1}\,}

\usepackage{fontawesome5}

\usepackage[most]{tcolorbox}
\usepackage{ragged2e}
\definecolor{promptBoundary}{RGB}{144, 182, 234}
\definecolor{promptBackground}{RGB}{234, 245, 253}
\newtcolorbox{myPromptQuote}{ enhanced, breakable, colback=promptBackground!40, colframe=promptBoundary!80, left=6pt, right=6pt, top=4pt, bottom=4pt, boxrule=0.6pt, arc=1mm, fontupper=\footnotesize,
before upper={\justifying\setlength{\parindent}{0pt}\setlength{\emergencystretch}{2em}},
}

\DeclareRobustCommand{\orcidicon}{
\begin{tikzpicture}
\draw[lime, fill=lime] (0,0)
circle[radius=0.16]
node[white]{{\fontfamily{qag}\selectfont \tiny \.{I}D}};
\end{tikzpicture}
\hspace{-2mm}
}
\foreach \x in {A, ..., Z}{%
\expandafter\xdef\csname orcid\x\endcsname{\noexpand\href{https://orcid.org/\csname orcidauthor\x\endcsname}{\noexpand\orcidicon}}
}

\begin{document}

\title{ARTEMIS: Agent-guided Reliability-aware Temporal Mask Evolution for Imperfectly Supervised Video Polyp Segmentation}

\author{
Tong Wang\hspace{-1.5mm}\orcidA{},
Siwen Wang\hspace{-1.5mm}\orcidD{},
Yaolei Qi\hspace{-1.5mm}\orcidE{},
Jinxing Zhou\hspace{-1.5mm}\orcidI{},
Yuting He\hspace{-1.5mm}\orcidJ{},
Guanyu Yang\hspace{-1.5mm}\orcidB{} and Yutong Xie\hspace{-1.5mm}\orcidC{}
\thanks{Tong Wang, Yaolei Qi, and Guanyu Yang are with the Key Laboratory of New Generation Artificial Intelligence Technology and Its Interdisciplinary Applications, Southeast University, Ministry of Education, Jiangsu, China. (email: tongwangnj@qq.com, yang.list@seu.edu.cn)}
\thanks{Siwen Wang, Jinxing Zhou, and Yutong Xie are with the Mohamed bin Zayed University of Artificial Intelligence (MBZUAI), Abu Dhabi, UAE. (email: yutong.xie678@gmail.com)}
\thanks{Yuting He is with the School of Medicine, Case Western Reserve University, Cleveland, OH, USA. (e-mail: yuting.he4@case.edu)}
\thanks{Corresponding authors: \emph{Guanyu~Yang} and \emph{Yutong~Xie}. Work was done when Tong Wang was visiting MBZUAI.}
}

\markboth{IEEE Transactions on Image Processing}%
{Wang \MakeLowercase{\textit{et al.}}: ARTEMIS for Imperfectly Supervised Video Polyp Segmentation}


\maketitle

\begin{abstract}
Imperfectly supervised video polyp segmentation (VPS) seeks to learn dense and temporally consistent masks from inexpensive supervision, covering weak annotations such as points and scribbles, as well as semi-supervision with only a small subset of densely labeled frames.
This setting is clinically attractive but technically challenging: polyps often suffer from weak object-background contrast, ambiguous boundaries, motion blur, and specular highlights, while incomplete supervision provides only sparse or missing pixel-level guidance.
Although SAM2 can transform sparse or partial annotations into dense masks, direct pseudo labeling still suffers from geometry-degraded pseudo masks with boundary leakage into background mucosa, temporally underused pseudo labeling, and reliability-blind pseudo supervision.
To address these issues, we propose \textbf{ARTEMIS}, a unified framework for imperfectly supervised VPS driven by agent-guided reliability-aware temporal mask evolution. ARTEMIS first initializes coarse masks from available supervision: points/scribbles are converted by SAM2, while dense labels in semi-supervision serve as reliable anchors. It then uses a debate-and-judge vision-language agent to select reliable temporal anchors under weak supervision and propagates them bidirectionally with SAM2 to refine unreliable or unlabeled frames. Finally, ARTEMIS trains the final segmenter with temporal reliability-aware robust learning, comprising reliability-guided reference selection, a Reference Prototype Transport Module, and reliability-aware robust loss. These components assess mask reliability, evolve anchors over time, transport target identity across frames, and down-weight noisy supervision rather than discarding difficult samples. Experiments on SUN-SEG and CVC-ClinicDB-612 under scribble, point, and limited-label settings show that ARTEMIS achieves state-of-the-art performance. 
Code will be released at \url{https://github.com/wangtong627/ARTEMIS}.
\end{abstract}


\section{Introduction}

\IEEEPARstart{C}{olorectal} cancer is the third most common cancer worldwide~\cite{bray2024global} and usually develops from polyps on the colonic mucosa~\cite{bernal2017comparative}. Although colonoscopy is the gold standard for screening, more than 25\% of polyps are missed~\cite{shaban2020context}, necessitating accurate and efficient automatic polyp segmentation.

\begin{figure}
    \centering
    \includegraphics[width=\linewidth]{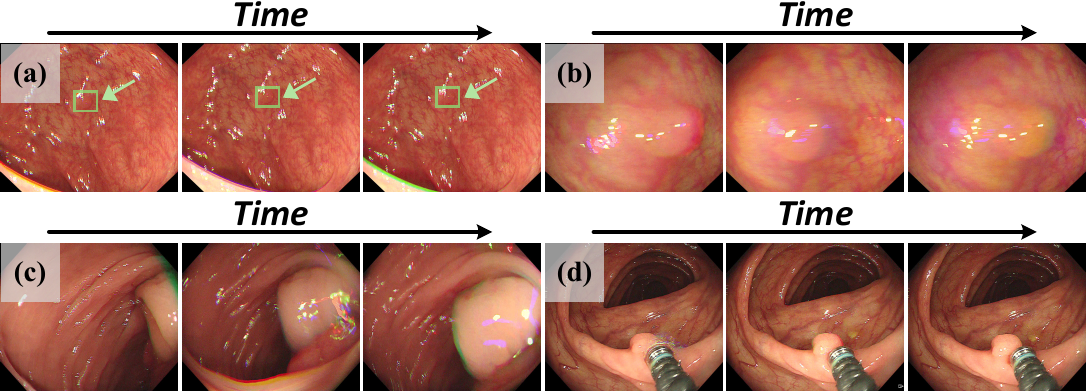}
    \vspace{-8mm}
    \caption{\textbf{Challenges in VPS.} (a) Low contrast between the foreground polyp within the green box and surrounding background; (b) motion blur; (c) scale variations; and (d) partial occlusion.}
    \vspace{-7mm}
    \label{fig:introduction_1}
\end{figure}

Video polyp segmentation (VPS)~\cite{ji2021progressively,ji2022video,hu2024sali,chen2025stddnet,wang2026cmsa} provides frame-level localization and tracking cues for computer-aided colonoscopy. Compared with static image segmentation~\cite{fan2020pranet,wang2024polyp,wu2022polypseg+,wu2025epsegnet,xiao2024ctnet,liu2024devil}, VPS is more challenging because polyps often share similar color, texture, and boundaries with surrounding mucosa. As shown in~\cref{fig:introduction_1}, this weak discriminability is further amplified by common clinical artifacts, including \textbf{1) background ambiguity}, \textbf{2) motion blur}, \textbf{3) scale variation}, and \textbf{4) partial occlusion}. These spatial-temporal perturbations often cause missed regions, over-segmentation, or temporal drift.

Supervision is another bottleneck. Dense annotation for colonoscopy videos is labor-intensive and costly: at 30 fps, a 30-second video has 900 frames and requires 75 hours at 5 minutes/frame~\cite{ji2021progressively}, making pixel-perfect masks impractical at scale. Lower-cost alternatives include weak supervision, such as point prompts ($\sim$2 seconds/frame) or scribbles ($\sim$10 seconds/frame)~\cite{yu2021structure,he2023weakly}, and semi-supervision with dense masks for only a small fraction of frames.
Existing methods address these regimes separately: \textbf{1) weakly supervised methods} rely on feature transformation, annotation expansion, or loss design with sparse labels~\cite{yu2021structure,liang2022tree,he2023weakly}; \textbf{2) semi-supervised methods} use teacher-student consistency or mutual correction~\cite{tarvainen2017mean,wang2023mcf,miao2023caussl,wu2024cross,zhao2024alternate}; and \textbf{3) unified video learning under imperfect supervision} remains underexplored.
The first two address individual regimes but do not unify imperfectly supervised VPS; they remain supervision-limited, lack priors for dense-label recovery, and provide limited temporal consistency.
Because colonoscopy datasets often mix sparse labels with partially labeled videos under different budgets, imperfect supervision is heterogeneous.
This motivates a \textit{unified setting for imperfectly supervised VPS} that handles both cases in one formulation, improving generality, reducing weak/semi-supervised fragmentation, and matching mixed supervision.
%

Foundation models offer a promising way to inject external visual priors into imperfect supervision. SAM~\cite{kirillov2023segment} and SAM2~\cite{ravi2024sam2} can generate dense masks from sparse prompts and have been used for weakly or incompletely supervised segmentation~\cite{he2023weakly_sam,zhao2025weakpolyp,he2025segment}. However, simply replacing missing labels with SAM-generated pseudo masks does not provide a unified solution for imperfectly supervised VPS, where pseudo labels must be both geometrically plausible and temporally coherent across supervision regimes. The limitations are threefold:
\textbf{1) Geometry-degraded pseudo masks.} Masks from points, scribbles, or initial models can be unreliable in low-contrast, blurred, fast-motion, and occluded frames, causing structural distortion, fragmented interiors, and boundary leakage into background mucosa~\cite{yu2021structure, liang2022tree, he2023weakly}; such failures are especially relevant to polyp segmentation with ambiguous boundaries;
\textbf{2) Temporally underused pseudo labeling.} Existing SAM-based methods for imperfect supervision mostly target image-level tasks or a single supervision regime, often performing one-shot pseudo-label generation or selection~\cite{he2023weakly_sam,chen2024sam,ge2026d3etor}; thus, they underuse SAM2's temporal memory and mask propagation for cross-frame continuity, which is central to VPS; and
\textbf{3) Reliability-blind pseudo supervision.} Heuristic filtering with area, confidence, or fixed thresholds may discard difficult but informative frames or retain noisy masks as equal supervision~\cite{hu2024relax,he2025segment,chen2024sam,chen2023sam,hu2025st,huang2025learnable}, both problematic for pseudo-label and noisy-label medical segmentation. These observations suggest that foundation-model masks should not be directly accepted or rejected as static labels; instead, they should be assessed for reliability, evolved through temporal propagation, and used with noise-aware supervision.

\begin{figure}
    \centering
    \includegraphics[width=\linewidth]{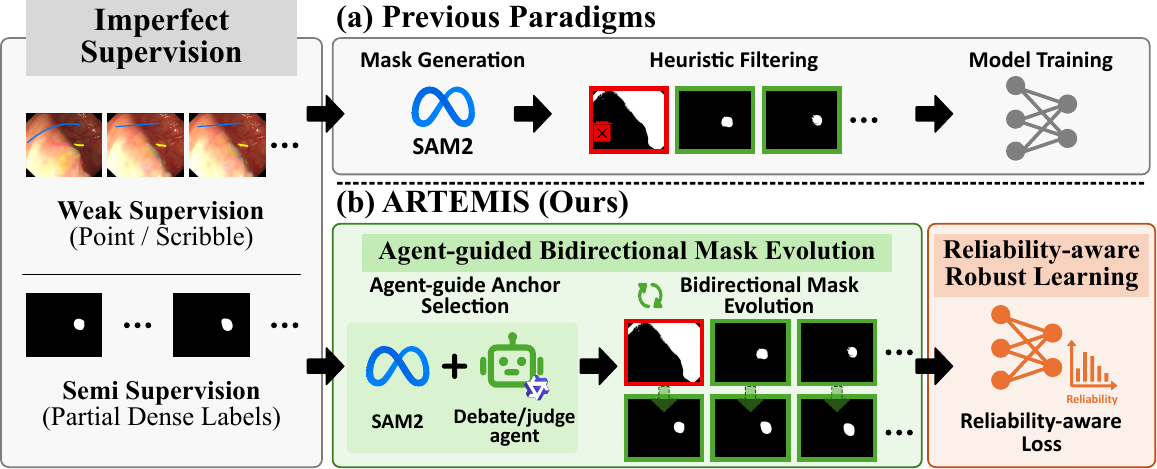}
    \vspace{-8mm}
    \caption{
    \textbf{Comparison of imperfectly supervised VPS paradigms.}
    Prior pipelines hard-filter pseudo labels, discarding useful samples and remaining noise-sensitive. ARTEMIS follows \textit{complete-then-learn}: Stage 1 evolves reliable anchors, while Stage 2 uses reliability-guided reference selection, RPTM, and robust loss across weak/semi-supervised settings.
    }
    \vspace{-7mm}
    \label{fig:intro_paradigm}
\end{figure}

To address these limitations, we propose \textbf{ARTEMIS}, an agent-guided reliability-aware temporal mask evolution framework for \textit{unified imperfectly supervised VPS}. ARTEMIS provides a unified framework for this underexplored setting by jointly covering weak supervision and semi-supervision, rather than treating them as isolated problems. As shown in \cref{fig:intro_paradigm}, ARTEMIS differs from prior paradigms in three aspects. \textbf{1)} It uses SAM2 and a debate-and-judge vision-language agent to identify reliable temporal anchors, injecting foundation-model priors into incomplete supervision. \textbf{2)} It evolves masks bidirectionally so reliable anchors refine unreliable or unlabeled frames through temporal propagation. \textbf{3)} It trains the final segmenter with reliability-guided reference selection, Reference Prototype Transport Module (RPTM), and reliability-aware robust loss, using informative imperfect pseudo labels instead of simply discarding them. For weak supervision, sparse scribbles/points are converted into dense masks; for semi-supervision, available dense masks serve as reliable anchors and are propagated to unlabeled frames.

Our main contributions are summarized as follows:

\begin{itemize}
\item \textbf{Unified Imperfectly Supervised VPS.} We establish a unified paradigm and propose ARTEMIS, a two-stage framework that handles weakly supervised and semi-supervised VPS by completing sparse or missing annotations into temporally consistent dense pseudo masks.

\item \textbf{Agent-guided bidirectional mask evolution.} We introduce a debate-and-judge vision-language agent to select reliable masks as weak-supervision anchors, while dense labels serve as semi-supervision anchors; both are bidirectionally propagated to refine unreliable frames.

\item \textbf{Temporal reliability-aware robust learning.} We combine reliability-guided reference selection, RPTM, and reliability-aware robust loss to enhance temporal identity consistency and suppress noisy supervision.

\item \textbf{Comprehensive validation.} Experiments under weakly supervised and semi-supervised settings on two benchmark datasets show state-of-the-art performance.
\end{itemize}

\section{Related Work}
\minisection{Video polyp segmentation.}
Early polyp segmentation methods~\cite{fan2020pranet, wang2024polyp, wu2022polypseg+, wu2025epsegnet, xiao2024ctnet, liu2024devil, sun2026asgnet, lu2024domain} mainly follow static image paradigms and perform well on image benchmarks, but directly applying them to videos ignores temporal correlations and may yield discontinuous masks.
To exploit video cues, PNS~\cite{ji2021progressively} and PNS+~\cite{ji2022video} introduce progressive temporal modeling, while recent VPS methods~\cite{hu2024sali,wang2026cmsa,chen2025stddnet,zhangwavepolyp,hehfsti,fang2024embedding} further study short-term alignment, long-term interaction, and Mamba-based dynamics. These works are mostly fully supervised, whereas ARTEMIS uses temporal modeling to complete sparse or missing annotations into dense pseudo masks.

\minisection{Imperfectly supervised segmentation.}
Weakly supervised segmentation~\cite{yu2021structure,liang2022tree,he2023weakly,hu2024relax,hu2024leveraging,zhao2025weakpolyp,zhou2021group,du2024pixel,zhang2023weakly,zhu2026weaktr,zhao2022semi,li2025boost,chen2026vpsentry} reduces annotation cost by learning from points, scribbles, or other sparse labels, usually with structural priors, local consistency, or task-specific losses.
Semi-supervised segmentation~\cite{tarvainen2017mean,wang2023mcf,miao2023caussl,wu2024cross,zhao2024alternate,huang2025uncertainty,wang2024boundary,cao2025background,niu2024minet,gao2025progressive} uses limited dense labels with teacher-student consistency, mutual correction, or cross-view learning. SEE~\cite{he2025segment} unifies weak and semi-supervised concealed object segmentation with SAM-based pseudo-label generation and selection. Yet most imperfectly supervised methods remain image-oriented or closed-loop, so they underuse temporal propagation; ARTEMIS instead formulates weak/semi-supervised VPS as bidirectional temporal pseudo-label completion.

\minisection{Foundation-model-assisted pseudo labeling.}
SAM~\cite{kirillov2023segment} and SAM2~\cite{ravi2024sam2} provide strong promptable mask generation, and SAM-based pseudo labels have been used for weakly supervised concealed or polyp segmentation~\cite{he2023weakly_sam,zhao2025weakpolyp} and incomplete supervision~\cite{he2025segment}. However, such masks remain unreliable in camouflaged or medical scenes with weak contrast and ambiguous boundaries. Structured conditional reasoning~\cite{shen2024imagpose,shen2024advancing} and multi-agent debate (MAD)~\cite{chan2308chateval,liang2024encouraging,qian2025scaling} offer complementary evidence and judge-based decisions, making them suitable for pseudo-label reliability assessment. Recent debate-enhanced pseudo-labeling applies this idea to weakly supervised camouflaged object detection~\cite{ge2026d3etor}; ARTEMIS adapts it to VPS by selecting reliable temporal anchors, evolving them with SAM2, and suppressing residual noise via our reliability-aware learning method.

\section{Methodology}
\subsection{Methodology Overview}
\begin{figure*}
    \centering
    \includegraphics[width=\linewidth]{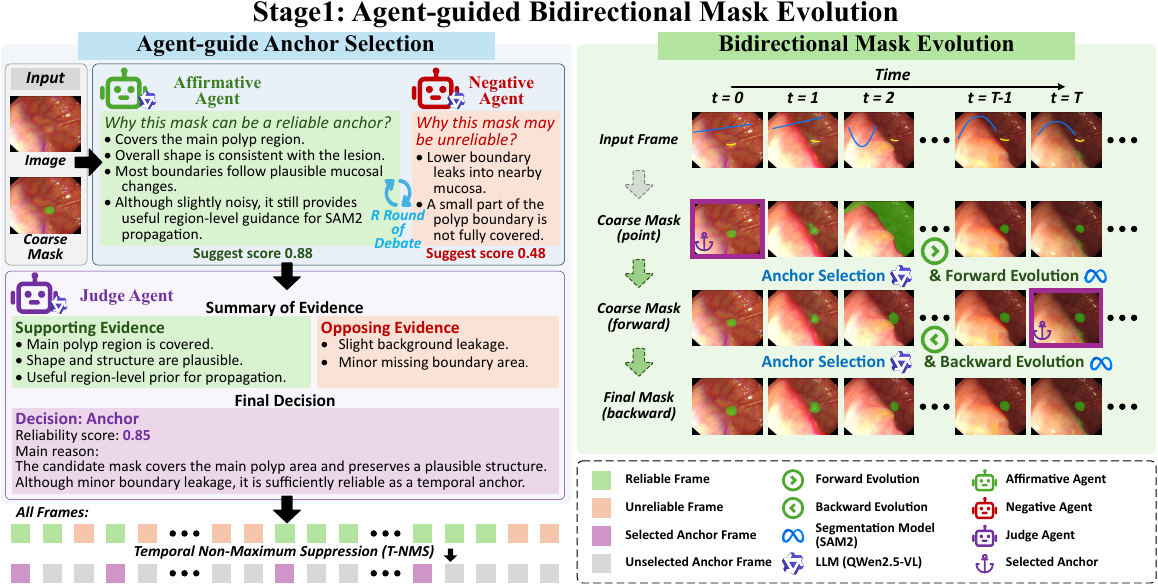}
    \vspace{-8mm}
    \caption{
    \textbf{Overview of Stage 1: Agent-guided Bidirectional Mask Evolution.}
    Given an input frame and SAM2 coarse mask, a debate-and-judge agent scores its anchor reliability.
    Candidate anchors are filtered by temporal NMS and propagated bidirectionally with SAM2 to produce evolved pseudo masks.
    }
    \vspace{-4mm}
    \label{fig:figure_method_stage_1}
\end{figure*}

\begin{figure*}
    \centering
    \includegraphics[width=\linewidth]{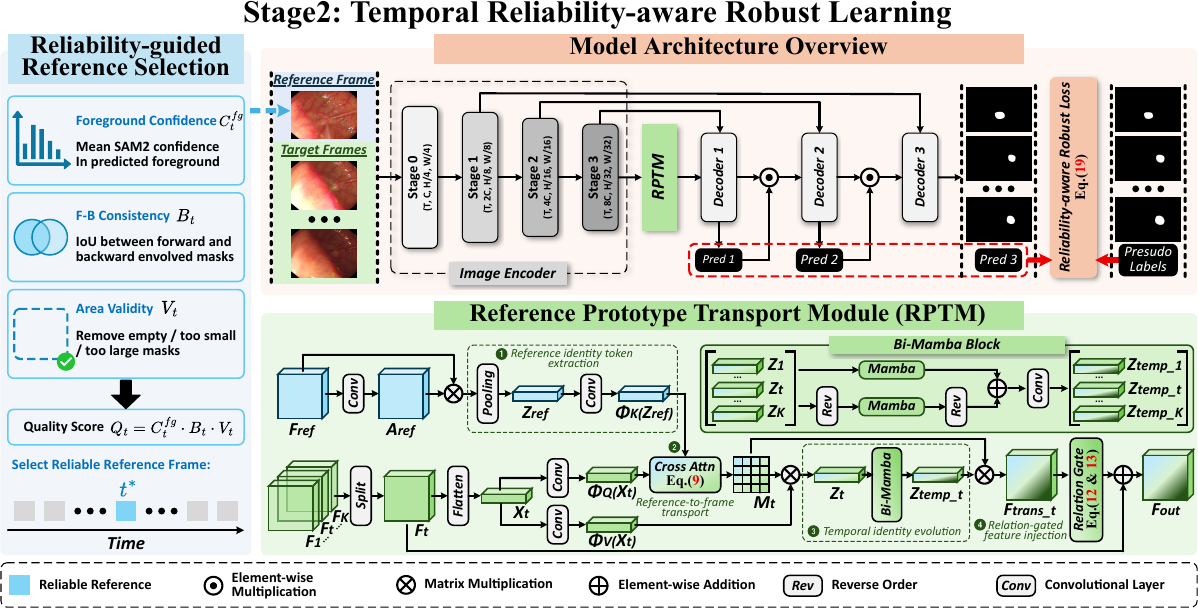}
    \vspace{-8mm}
    \caption{
    \textbf{Overview of Stage 2: Temporal Reliability-aware Robust Learning.}
    ARTEMIS selects a reliable reference using mask reliability cues, transports reference identity tokens to target frames via RPTM, and applies a reliability-aware robust loss to suppress residual pseudo-label noise.
    }
    \vspace{-4mm}
    \label{fig:figure_method_stage_2}
\end{figure*}

This work follows a \textit{complete-then-learn} pipeline. \textbf{Stage 1} converts available supervision into coarse masks, selects reliable temporal anchors, and performs SAM2-based bidirectional propagation to obtain evolved pseudo masks. Weak supervision requires scribble/point-to-mask completion, while semi-supervision directly uses dense labels as anchors. \textbf{Stage 2} trains the segmenter with reliability-guided reference selection, RPTM, and reliability-aware robust loss, see \cref{fig:figure_method_stage_1} and \cref{fig:figure_method_stage_2}.

\subsection{Motivation for Temporal Mask Evolution}

\begin{table}[t]
\centering
\caption{
Comparison of different prompt types on polyp segmentation.
}
\label{tab:observation1_prompt_observation}
\vspace{-2mm}
\setlength{\tabcolsep}{6pt}
\renewcommand{\arraystretch}{0.95}
\resizebox{\columnwidth}{!}{
\begin{tabular}{lcc}
\toprule
\textbf{Prompt Type} & \textbf{Mean Dice (\%) $\uparrow$} & \textbf{Mean MAE (\%) $\downarrow$} \\
\midrule
Sparse Point
& 71.8
& 13.7 \\
Light Noisy Mask ($\beta=0.2$)
& 88.3 {\scriptsize $(+23.0\%)$}
& 2.3 {\scriptsize $(-83.4\%)$} \\
Moderate Noisy Mask ($\beta=0.4$)
& 81.3 {\scriptsize $(+13.2\%)$}
& 3.0 {\scriptsize $(-78.4\%)$} \\
Heavy Noisy Mask ($\beta=0.6$)
& 74.3 {\scriptsize $(+3.4\%)$}
& 3.1 {\scriptsize $(-77.2\%)$} \\
\bottomrule
\end{tabular}
}
\end{table}

\begin{figure}[t]
\vspace{-2mm}
    \centering
    \includegraphics[width=\linewidth]{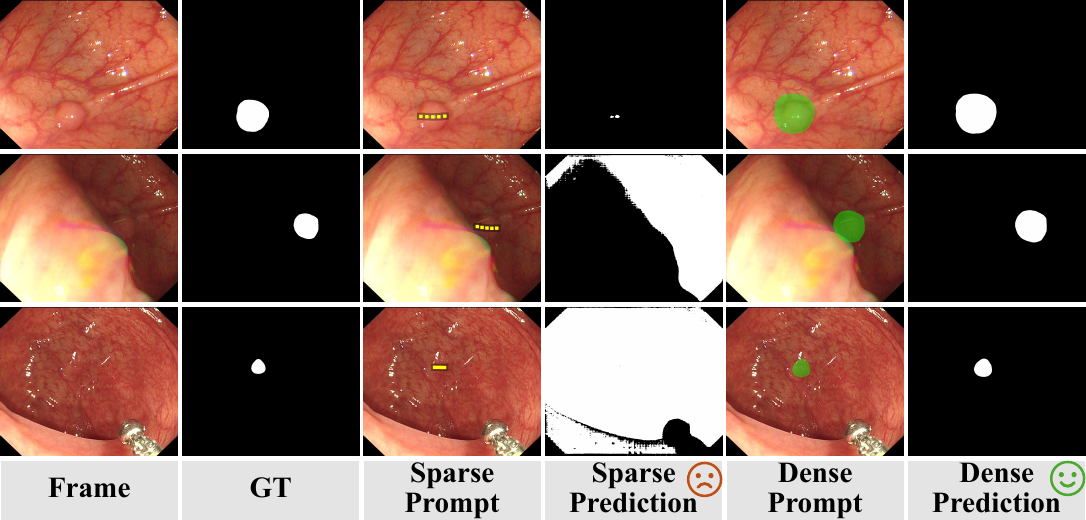}
    \vspace{-8mm}
    \caption{
    \textbf{Qualitative comparison of sparse and dense prompts.}
    Sparse Prompt uses yellow point prompts, while Dense Prompt uses green heavy noisy mask prompts on the same frames.
    Compared with sparse prompts, dense noisy masks provide stronger spatial priors for polyp localization.
    }
    \label{fig:observation1}
    \vspace{-4mm}
\end{figure}

\minisection{{\color{orange}\faLightbulb}~Observation 1: Dense mask prompt is more informative for weakly discriminative targets.}
A key challenge in VPS is the weak semantic discriminability. Their color, texture, and boundaries can be highly similar to surrounding mucosa, making it difficult to infer complete object extent from local cues alone.
Sparse point prompts provide only discrete locations and cannot sufficiently describe target shape, boundary, or spatial extent. In contrast, dense mask prompts offer region-level priors and stronger structural constraints. \textbf{\textit{This motivates us to ask: for weakly discriminative polyps, are dense mask prompts more effective than sparse points, even when noisy?}}

We conduct an offline observation experiment on a SUN-SEG subset with 10 colonoscopy videos and 1,478 matched frames. Ground-truth masks are used only for this diagnostic study to synthesize controlled noisy masks and compute Dice/MAE, not for ARTEMIS training. Sparse points sampled from scribbles~\cite{he2025segment} serve as the baseline. Following~\cite{li2021superpixel}, noisy dense masks are generated from ground truth using random dilation/erosion and affine perturbation. The perturbation ratio $\beta$ denotes the changed-pixel proportion, with $\beta=0.2$, $0.4$, and $0.6$ for light, moderate, and heavy noise.

As shown in~\cref{tab:observation1_prompt_observation}, noisy mask prompts consistently outperform sparse points, even under heavy perturbation. This indicates that approximate region support is more informative than isolated point cues for weakly discriminative polyps.
As illustrated in~\cref{fig:observation1}, yellow sparse points often cause under- or over-segmentation, whereas green heavy noisy masks provide stronger region-level guidance. This observation leads to a key design choice: ARTEMIS uses sparse prompts only as low-cost localization cues and converts them into coarse dense masks for subsequent \textit{selection}, \textit{propagation}, and \textit{refinement}.

\begin{table}[t]
\centering
\caption{
Comparison of prompt types for polyp segmentation.
For noisy masks, frames with Dice $<0.6$ are replaced by dense noisy masks and propagated by SAM2 to the next five frames without additional prompts.
}
\label{tab:prompt_observation_dense_noisy_mask}
\vspace{-2mm}
\setlength{\tabcolsep}{6pt}
\renewcommand{\arraystretch}{0.95}
\resizebox{\columnwidth}{!}{
\begin{tabular}{lcc}
\toprule
\textbf{Prompt Type}
& \textbf{Mean Dice (\%) $\uparrow$}
& \textbf{Mean MAE (\%) $\downarrow$} \\
\midrule
Sparse Point
& 71.8
& 13.7 \\
Light Noisy Mask ($\beta=0.2$)
& 92.4 {\scriptsize $(+28.6\%)$}
& 0.7 {\scriptsize $(-95.1\%)$} \\
Moderate Noisy Mask ($\beta=0.4$)
& 89.9 {\scriptsize $(+25.2\%)$}
& 0.7 {\scriptsize $(-94.7\%)$} \\
Heavy Noisy Mask ($\beta=0.6$)
& 86.7 {\scriptsize $(+20.8\%)$}
& 0.8 {\scriptsize $(-94.4\%)$} \\
\bottomrule
\end{tabular}
}
\end{table}

\begin{figure}
\vspace{-3.5mm}
    \centering
    \includegraphics[width=\linewidth]{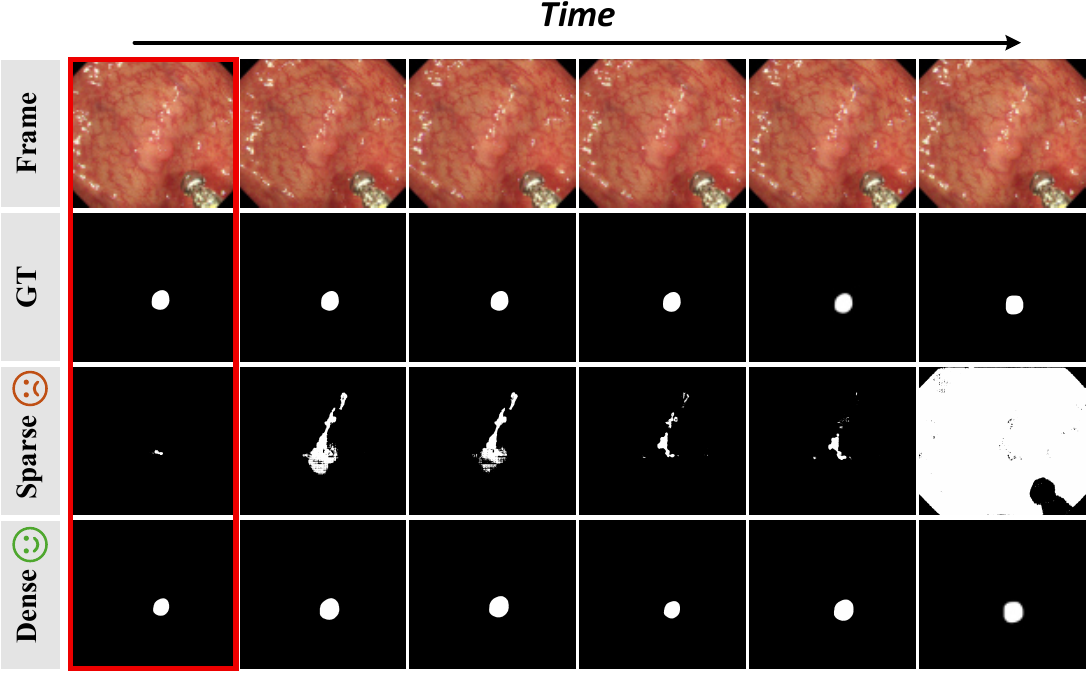}
    \vspace{-8mm}
    \caption{
    \textbf{Qualitative results of temporal propagation from reliable anchors.}
    The red box marks the injected dense mask prompt, which improves subsequent frames without additional prompts.
}
    \label{fig:observation2}
    \vspace{-4mm}
\end{figure}

\minisection{{\color{orange}\faLightbulb}~Observation 2: High-quality mask prompts benefit subsequent frames through temporal propagation.}
In SAM2 video inference, a prompted frame is encoded into the temporal memory bank and can influence later frames. A reliable mask prompt may therefore act as a temporal anchor rather than only a single-frame correction. \textbf{\textit{This motivates us to ask: if a dense mask prompt is injected into a difficult frame, can its benefit extend to the following frames?}}

We use sparse point prompting as the baseline and identify low-quality frames by Dice $<0.6$. This criterion is used only in the observation study with ground truth, not as a training-time rule. For each selected frame, we replace the point prompt with a dense noisy mask in Observation 1, and let SAM2 propagate to the next five frames without additional prompts.

As shown in~\cref{tab:prompt_observation_dense_noisy_mask}, dense noisy masks consistently improve temporal propagation. Dice increases from 71.8\% to 92.4\%, 89.9\%, and 86.7\% under light, moderate, and heavy noise, while MAE decreases from 13.7\% to below 0.8\%.
The qualitative results in~\cref{fig:observation2} confirm this temporal effect: replacing a poor point prompt with a dense noisy mask refines the prompted frame and improves the following propagated frames. Since this observation only verifies forward propagation, ARTEMIS further applies the same mechanism in the reverse temporal direction. This yields \textit{bidirectional mask evolution}, where reliable anchors refine low-quality frames both before and after them.

\subsection{Stage 1: Agent-guided Bidirectional Mask Evolution}

\minisection{From observations to design.}
Given a video $\{I_t\}_{t=1}^{T}$ with $I_t\in\mathbb{R}^{3\times H\times W}$, where $t$ is the frame index, $T$ is the frame number, and $H,W$ are frame height and width, ARTEMIS first obtains an initial coarse mask $M_t^c\in[0,1]^{H\times W}$ from the available supervision. The supervision source determines how anchors are formed: \textbf{1)} for weak supervision, scribbles are converted into sparse points and point annotations are used directly; SAM2 then converts these sparse prompts into coarse dense masks. \textbf{2)} For semi-supervision, available dense labels are directly used as mask prompts. Stage 1 then selects or defines reliable masks as anchors and propagates them bidirectionally to refine unlabeled or unreliable frames.

\minisection{Agent-guided reliable anchor selection.}
Directly propagating all coarse masks may contaminate the SAM2 memory bank, so ARTEMIS first estimates which masks are reliable enough to serve as anchors. For each image-mask pair $(I_t, M_t^c)$, we provide the original image, candidate mask, mask overlay, and a fixed instruction to a debate-and-judge vision-language agent. We instantiate the agent with Qwen2.5-VL-7B~\cite{Qwen2.5-VL}. The agent contains three roles: \textbf{1) affirmative agent}, which argues for accepting the mask; \textbf{2) negative agent}, which identifies failure risks; and \textbf{3) judge agent}, which outputs an anchor/non-anchor decision with a reliability score $s_t \in [0,1]$, clipped to this range for normalization.

The fixed prompt used for all frames is:

\begin{myPromptQuote}
\textbf{Meta Prompt.}
You are an expert agent for colonoscopy video polyp segmentation.
Given an image and a candidate coarse mask generated by SAM2, evaluate whether the mask is reliable enough to serve as a temporal anchor for mask propagation.
Focus on image-mask alignment, foreground completeness, boundary plausibility, background leakage, artifact confusion, and overall credibility as a mask prompt.

\vspace{2mm}

\textbf{Affirmative Agent.}
Argue why the candidate mask should be accepted as a reliable anchor.
Identify visual evidence showing that it covers the main polyp region, preserves plausible structure, and can provide useful region-level guidance for SAM2 propagation.

\vspace{2mm}

\textbf{Negative Agent.}
Argue why the candidate mask may be unreliable.
Identify visual evidence of under-segmentation, over-segmentation, boundary errors, background leakage, or artifact confusion, and explain possible risks if it is propagated.

\vspace{2mm}

\textbf{Judge Agent.}
Given the image, candidate mask, and debate history, summarize the key evidence from both sides.
Output: Decision: Anchor / Non-anchor; Reliability score: 0--1; Main reason: ...; Failure risk: ...
\end{myPromptQuote}

Formally, let $D^{+}$ and $D^{-}$ denote the affirmative and negative debating agents, respectively, and let $P_{\text{meta}}$ denote the shared meta prompt. For each frame $t$, the two agents produce supporting and opposing assessments:
\begin{equation}
a_t^{+} = D^{+}(I_t, M_t^c, P_{\text{meta}}),
\quad
a_t^{-} = D^{-}(I_t, M_t^c, P_{\text{meta}}),
\end{equation}
where $a_t^{+}$ and $a_t^{-}$ denote the positive and negative arguments for the candidate mask $M_t^c$. After $R$ debate rounds, the judge agent $J$ takes the complete debate history $\mathcal{H}_t$ as input and outputs the final reliability score $s_t$:
\begin{equation}
s_t = J(I_t, M_t^c, \mathcal{H}_t).
\end{equation}

We first retain frames whose reliability scores exceed the anchor selection threshold $\tau_a$. In this paper, we set $\tau_a=0.7$. The resulting candidate anchor set $\mathcal{C}$ is defined as:
\begin{equation}
\mathcal{C} = \{t \mid s_t > \tau_a\}.
\end{equation}

To improve temporal coverage, we further apply temporal non-maximum suppression (T-NMS): candidates in $\mathcal{C}$ are sorted by reliability scores $\{s_t\}_{t=1}^{T}$, and lower-scoring candidates within a radius of $r=5$ frames around a selected anchor are suppressed. The final anchor set $\mathcal{A}$ is:
\begin{equation}
\mathcal{A} = \operatorname{T\mbox{-}NMS}(\mathcal{C}, \{s_t\}_{t=1}^{T}, r).
\end{equation}

If no frame satisfies $s_t>\tau_a$, we select the highest-scoring frame as a fallback anchor to avoid unprompted propagation. Selected anchors are injected into SAM2 as mask prompts, while non-anchor frames are refined by bidirectional mask evolution.
The following example illustrates the process.

\begin{myPromptQuote}

\textbf{Input.}
Image--mask pair $(I_t, M_t^c)$, where the candidate coarse mask roughly covers the main polyp region but has slight boundary leakage.

\vspace{2mm}

\textbf{Affirmative Agent.}
The candidate mask should be accepted as a reliable anchor.
It covers the main polyp region, preserves a plausible object shape, and provides useful region-level guidance for SAM2 temporal propagation.
Although the boundary is slightly noisy, the main target structure is still retained.
Suggested score: 0.88.

\vspace{2mm}

\textbf{Negative Agent.}
The candidate mask may contain several risks.
The lower boundary slightly leaks into nearby mucosa, and a small part of the polyp boundary is not fully covered.
If propagated directly, this leakage may affect nearby frames.
Suggested score: 0.48.

\vspace{2mm}

\textbf{Debate Refinement.}
The affirmative agent argues that the leakage is local and does not dominate the mask, while the negative agent agrees that the main failure is mild.
Both agents indicate that the mask is imperfect but more reliable than sparse point prediction.

\vspace{2mm}

\textbf{Judge Agent.}
Decision: Anchor.
Reliability score: 0.85.
Main reason: the mask covers the main polyp area and preserves a plausible structure.
Failure risk: small background leakage may be propagated, but the risk is acceptable.

\vspace{2mm}

\textbf{Output.}
Since $s_t=0.85>\tau_a$, this frame is selected as a reliable temporal anchor.
\end{myPromptQuote}

\minisection{Bidirectional mask evolution.}
Given anchors $\mathcal{A}$, ARTEMIS injects each anchor mask $M_t^c$ into SAM2 as a prompt and propagates it. A forward pass refines later frames but cannot correct earlier ones. Therefore, ARTEMIS performs a second pass in reverse order, allowing reliable anchors to influence both past and future frames. Let $L_t^f,L_t^b\in\mathbb{R}^{H\times W}$ be forward/backward SAM2 logits and $P_t^f=\sigma(L_t^f)$, $P_t^b=\sigma(L_t^b)$ be their probability maps, where $\sigma(\cdot)$ is sigmoid. The backward result is used as the evolved pseudo mask $\tilde{Y}_t\in[0,1]^{H\times W}$, while $P_t^f$ and $P_t^b$ are for reliability estimation.

\minisection{Adaptation to semi-supervision.}
In the semi-supervised setting, the sparsely available dense labels are already high-quality mask prompts. Therefore, we treat all labeled frames as reliable anchors, set their reliability scores to $s_t=1$, and directly propagate these dense masks forward and backward with SAM2 to complete pseudo labels for unlabeled frames.

\subsection{Stage 2: Temporal Reliability-aware Robust Learning}

Although Stage 1 improves pseudo masks, the evolved masks may still contain boundary errors, leakage, or incomplete regions. Stage 2 is designed for this residual noise: it selects a reliable reference frame, transports target identity across time, and down-weights unreliable supervision.

\minisection{Reliability-guided reference selection.}
Under pseudo-label supervision, a fixed first-frame reference may be unreliable due to motion blur, weak visibility, or inaccurate masks. ARTEMIS therefore selects a reliability-guided reference frame from evolved pseudo masks.
For each frame $t$, we define the frame-level quality score $Q_t\in[0,1]$ as:
\begin{equation}
\vspace{-1mm}
Q_t = C_t^{fg} \cdot B_t \cdot V_t,
\vspace{-1mm}
\end{equation}
where $C_t^{fg}\in[0,1]$ measures foreground confidence, $B_t\in[0,1]$ measures forward-backward agreement, and $V_t\in\{0,1\}$ is an area validity term, all computed from Stage 1 outputs without ground truth. Let $\Omega_t\subseteq\{1,\dots,H\}\times\{1,\dots,W\}$ be the foreground pixel set of the final backward pseudo mask and $|\Omega_t|$ its pixel count. The foreground confidence is:
\begin{equation}
\vspace{-1mm}
C_t^{fg}=\frac{1}{|\Omega_t|}\sum_{i\in\Omega_t}2|P_t^b(i)-0.5|,
\vspace{-1mm}
\end{equation}
where $i$ indexes a pixel location. We set $C_t^{fg}=0$ when $\Omega_t$ is empty. The agreement term $B_t$ is the IoU between forward- and backward-evolved masks. The validity term $V_t$ equals 1 only when the foreground area ratio lies in $[\rho_{min},\rho_{max}]$, where $\rho_{min}$ and $\rho_{max}$ are valid-area bounds. The reliable reference frame index $t_{ref}^{*}\in\{1,\dots,T\}$ is selected by:
\begin{equation}
\vspace{-1mm}
t_{ref}^{*} = \arg\max_t Q_t.
\vspace{-1mm}
\end{equation}
To reduce the discrepancy between this strategy and the common first-frame reference setting, the training reference index $t_{ref}\in\{1,\dots,T\}$ is sampled as $t_{ref}^{*}$ with probability $1-p$ and as the first frame with probability $p$, where $p$ is the first-frame sampling probability.

\minisection{Reference Prototype Transport Module (RPTM).}
The similarity between polyps and surrounding mucosa makes temporal modeling vulnerable to drift. RPTM uses a reliable reference frame as the identity source and transports its target identity to $K$ target frames. It is applied to the fused decoder feature before the prediction head, after encoder features have been mapped to a common channel dimension. The module contains four steps: \textbf{1)} reference identity token extraction; \textbf{2)} reference-to-frame transport; \textbf{3)} temporal identity evolution; and \textbf{4)} relation-gated feature injection. Let $F_{ref}\in\mathbb{R}^{C\times H_f\times W_f}$ and $F_t\in\mathbb{R}^{C\times H_f\times W_f}$ denote reference and target features, where $K$ is the number of target frames, $C$ is the channel number, $H_f,W_f$ are feature-map height and width, and subscript $f$ denotes feature resolution.

\textbf{1) Reference identity token extraction.}
RPTM first predicts $N$ spatial assignment maps $A_{ref}\in\mathbb{R}^{N\times H_f\times W_f}$ from $F_{ref}$, where $N$ is the number of identity tokens, and softly pools the reference feature:
\begin{equation}
\vspace{-1mm}
Z_{ref}^{n}
=
\sum_{h,w}
A_{ref}^{n}(h,w) F_{ref}(h,w),
\quad n=1,\dots,N,
\vspace{-1mm}
\end{equation}
where $h,w$ index feature-map locations and $Z_{ref}^{n} \in \mathbb{R}^{C}$ is the $n$-th token. $Z_{ref} \in \mathbb{R}^{N \times C}$ encodes reference target identity.

\textbf{2) Reference-to-frame transport.}
RPTM then matches reference tokens with dense target-frame locations. Let $X_t=\operatorname{Flatten}(F_t)\in\mathbb{R}^{H_fW_f\times C}$ be the flattened target feature. The transport map $M_t\in\mathbb{R}^{H_fW_f\times N}$ is:
\begin{equation}
\vspace{-1mm}
M_t =
\operatorname{Softmax}
\left(
\frac{\phi_q(X_t)\phi_k(Z_{ref})^{\top}}{\sqrt{C}}
\right),
\vspace{-1mm}
\end{equation}
where $\phi_q(X_t)\in\mathbb{R}^{H_fW_f\times C}$ and $\phi_k(Z_{ref})\in\mathbb{R}^{N\times C}$ are query and key projections, and softmax is applied over tokens. After normalizing $M_t$ over spatial locations to obtain $\bar{M}_t\in\mathbb{R}^{H_fW_f\times N}$, which acts as spatial attention maps for each token, target tokens are:
\begin{equation}
\vspace{-1mm}
Z_t = \bar{M}_t^{\top} \phi_v(X_t),
\vspace{-1mm}
\end{equation}
where $\phi_v(X_t)\in\mathbb{R}^{H_fW_f\times C}$ is the value projection and $Z_t\in\mathbb{R}^{N\times C}$. This reference-conditioned pooling suppresses visually similar but reference-inconsistent background regions.

\textbf{3) Temporal identity evolution.}
To efficiently model long-range temporal dependencies, we use bidirectional Mamba to evolve target identity tokens.
The target tokens are organized as $Z=[Z_1,\dots,Z_K]\in\mathbb{R}^{K\times N\times C}$:
\begin{equation}
\vspace{-1mm}
Z^{temp}
=
\phi_{fuse}
\left(
\operatorname{Mamba}_{f}(Z),
\operatorname{Rev}(\operatorname{Mamba}_{b}(\operatorname{Rev}(Z)))
\right),
\end{equation}
where $Z^{temp}\in\mathbb{R}^{K\times N\times C}$, $\operatorname{Mamba}_{f}$ and $\operatorname{Mamba}_{b}$ are forward/backward temporal blocks, $\operatorname{Rev}(\cdot)$ reverses temporal order, and $\phi_{fuse}(\cdot)$ fuses both directions.

\textbf{4) Relation-gated feature injection.}
For each target frame, $Z_t^{temp}\in\mathbb{R}^{N\times C}$ is transported back through $M_t$ as $F_t^{trans}=\operatorname{Reshape}(M_t Z_t^{temp})\in\mathbb{R}^{C\times H_f\times W_f}$, where $Z_t^{temp}$ is the $t$-th slice of $Z^{temp}$ and $\operatorname{Reshape}(\cdot)$ restores the $H_f\times W_f$ layout. A relation gate $G_t$ controls identity injection:
\begin{equation}
\vspace{-1mm}
G_t
=
\sigma
\left(
\phi_g
\left(
[F_t, F_t^{trans}, F_t-F_t^{trans}]
\right)
\right),
\end{equation}
where $[\cdot]$ denotes channel-wise concatenation and $\phi_g(\cdot)$ maps the $3C$-channel input to $G_t\in[0,1]^{C\times H_f\times W_f}$. The output is:
\begin{equation}
F_t^{out}
=
F_t
+
\gamma_m
\cdot
G_t
\odot
F_t^{trans},
\vspace{-1mm}
\end{equation}
where $F_t^{out}\in\mathbb{R}^{C\times H_f\times W_f}$, $\odot$ denotes element-wise multiplication, and $\gamma_m\in\mathbb{R}$ is a learnable scalar initialized to 0, making RPTM an identity mapping at the beginning.

\minisection{Reliability-aware robust loss.}
After temporal feature enhancement, ARTEMIS down-weights unreliable supervision at both \textit{pixel} and \textit{frame} levels.

For frame $t$, we define the final supervision weight $W_t\in[0,1]^{H\times W}$ as:
\begin{equation}
\vspace{-0.5mm}
W_t = R_t^{pix} \cdot w_t^{judge},
\end{equation}
where $R_t^{pix}\in[0,1]^{H\times W}$ is the pixel-level reliability map and $w_t^{judge}\in[0,1]$ is the frame-level judge weight. Given Stage 1 probability maps $P_t^{f}$ and $P_t^{b}$, confidence reliability $R_t^{conf}$ and soft bidirectional consistency $R_t^{bidir}$ are:
\vspace{-0.5mm}
\begin{equation}
R_t^{conf}=2|P_t^{b}-0.5|,
\quad
R_t^{bidir}=1-|P_t^{f}-P_t^{b}|,
\end{equation}
where $R_t^{conf},R_t^{bidir}\in[0,1]^{H\times W}$ and $|\cdot|$ is element-wise absolute value. The pixel-level reliability map is:
\begin{equation}
R_t^{pix}
=
\left(
R_t^{conf}
\cdot
R_t^{bidir}
\right)^{\gamma_r},
\end{equation}
where $\gamma_r>0$ controls the sharpness of reliability weighting. Under weak supervision, the frame-level weight $w_t^{judge}$ is assigned from the Stage 1 agent judgment: $w_t^{judge}=1.0$ if the frame is qualified as a reliable anchor with $s_t>\tau_a$, and $w_t^{judge}=0.5$ otherwise. Under semi-supervision, available dense masks are treated as reliable supervision.

Given the prediction $\hat{Y}_t\in[0,1]^{H\times W}$, evolved pseudo mask $\tilde{Y}_t\in[0,1]^{H\times W}$, and weight map $W_t$, we define $\mathcal{L}_{wbce}$ as:
\begin{equation}
\mathcal{L}_{wbce}
=
\frac{
\langle W_t, \operatorname{BCE}(\hat{Y}_t,\tilde{Y}_t) \rangle
}{
\langle W_t,\mathbf{1} \rangle + \epsilon
},
\end{equation}
and the weighted Dice loss as:
\begin{equation}
\mathcal{L}_{wdice}
=
1-
\frac{
2\langle W_t,\hat{Y}_t \odot \tilde{Y}_t\rangle+\epsilon
}{
\langle W_t,\hat{Y}_t\rangle+
\langle W_t,\tilde{Y}_t\rangle+\epsilon
}.
\vspace{-0.5mm}
\end{equation}
Here $\langle\cdot,\cdot\rangle$ denotes summation over pixels, $\mathbf{1}$ is an all-ones map, $\epsilon$ is a small constant for numerical stability, and $\operatorname{BCE}(\cdot)$ is pixel-wise binary cross-entropy.
The final loss $\mathcal{L}_{seg}$ is:
\begin{equation}
\mathcal{L}_{seg}
=
\lambda_{bce}\mathcal{L}_{wbce}
+
\lambda_{dice}\mathcal{L}_{wdice},
\vspace{-0.5mm}
\end{equation}
where $\lambda_{bce}$ and $\lambda_{dice}$ balance the two terms. During training, the decoder produces multi-scale side predictions for deep supervision; $\mathcal{L}_{seg}$ is computed for each output and averaged.

\subsection{Implementation Details}
We use SAM2~\cite{ravi2024sam2} (\texttt{sam2.1\_hiera\_base\_plus.pt}). The debate-and-judge agent is instantiated with Qwen2.5-VL-7B~\cite{Qwen2.5-VL} and uses one affirmative/negative debate round followed by one judge decision ($R=1$). Stage 1 uses anchor threshold $\tau_a=0.7$, T-NMS radius $r=5$. Stage 2 uses $K=5$ target frames, first-frame sampling probability $p=0.5$, area bounds $\rho_{min}=0.0005$ and $\rho_{max}=0.5$, reliability power $\gamma_r=2.0$, judge weights $w_t^{judge}=1.0/0.5$ for frames with/without $s_t>\tau_a$, and loss weights $\lambda_{bce}=0.5$, $\lambda_{dice}=1.0$. RPTM uses $N=4$ identity tokens, one forward and one backward Mamba block~\cite{mamba} with state dimension $d_{state}=16$, convolution width $d_{conv}=3$, expansion ratio 2, and relation gate. The segmentation backbone is PVT-v2-B2~\cite{Wang2021PVTVI}, and RPTM is applied to the 32-channel fused decoder~\cite{chen2025stddnet} feature map before the prediction head. The segmenter is trained using AdamW with batch size 12, weight decay $10^{-4}$, input size $352\times352$.
The base learning rate is $10^{-4}$ for decoder/head parameters and $10^{-5}$ for the backbone.
All experiments are conducted on one NVIDIA A100 GPU.

\section{Experiment}

\begin{table*}[t]
\centering
\caption{Quantitative comparison on the \textbf{SUN-SEG-Easy} and \textbf{SUN-SEG-Hard} datasets under \textit{Fully Supervised}, \textit{Scribble Supervision}, and \textit{Point Supervision} settings. The best and second-best results are highlighted in \textbf{bold} and \underline{underlined}, respectively.}
\vspace{-2mm}
\setlength{\tabcolsep}{0.005pt}
\renewcommand{\arraystretch}{0.93}
\resizebox{\textwidth}{!}{
\begin{tabular}{l|cccccc|cccccc|cccccc|cccccc}
\toprule
\multirow{2}{*}{Method} & \multicolumn{6}{c|}{SUN-SEG-Easy-Seen (\%)} & \multicolumn{6}{c|}{SUN-SEG-Easy-Unseen (\%)} & \multicolumn{6}{c|}{SUN-SEG-Hard-Seen (\%)} & \multicolumn{6}{c}{SUN-SEG-Hard-Unseen (\%)} \\
& $S_\alpha\uparrow$ & $E_{\phi}^{mn}\uparrow$ & $F_\beta^w\uparrow$ & Dice$\uparrow$ & IoU$\uparrow$ & MAE$\downarrow$ & $S_\alpha\uparrow$ & $E_{\phi}^{mn}\uparrow$ & $F_\beta^w\uparrow$ & Dice$\uparrow$ & IoU$\uparrow$ & MAE$\downarrow$ & $S_\alpha\uparrow$ & $E_{\phi}^{mn}\uparrow$ & $F_\beta^w\uparrow$ & Dice$\uparrow$ & IoU$\uparrow$ & MAE$\downarrow$ & $S_\alpha\uparrow$ & $E_{\phi}^{mn}\uparrow$ & $F_\beta^w\uparrow$ & Dice$\uparrow$ & IoU$\uparrow$ & MAE$\downarrow$ \\
\midrule

\multicolumn{25}{l}{\textit{Fully Supervised}} \\
\midrule
PraNet~\cite{fan2020pranet} & 91.8 & 94.2 & 87.7 & 88.3 & 82.5 & 2.0 & 78.1 & 78.8 & 66.3 & 66.5 & 58.2 & 5.2 & 88.4 & 91.9 & 83.1 & 83.9 & 76.6 & 3.1 & 78.7 & 80.2 & 66.7 & 67.5 & 58.7 & 5.3 \\
PNS~\cite{ji2021progressively} & 90.6 & 91.0 & 83.6 & 84.1 & 78.3 & 2.0 & 76.7 & 74.4 & 61.6 & 61.8 & 54.5 & 4.8 & 87.0 & 89.2 & 78.7 & 79.6 & 72.1 & 3.3 & 76.7 & 75.5 & 60.9 & 61.5 & 53.9 & 5.0 \\
PNS+~\cite{ji2022video} & 91.7 & 92.5 & 84.8 & 85.5 & 78.7 & 2.1 & 80.6 & 79.8 & 67.6 & 67.8 & 59.1 & 4.4 & 88.7 & 90.2 & 80.6 & 81.3 & 72.8 & 3.0 & 79.8 & 79.3 & 65.4 & 66.1 & 57.1 & 5.0 \\
\midrule

\multicolumn{25}{l}{\textit{Scribble Supervision}} \\
\midrule
SCWS~\cite{yu2021structure} & 79.3 & 83.4 & 50.3 & 74.5 & 65.1 & 10.0 & 73.6 & 74.2 & 44.5 & 60.3 & 50.4 & 13.5 & 76.6 & 80.7 & 47.5 & 70.8 & 61.2 & 12.3 & 72.2 & 74.7 & 41.9 & 60.4 & 50.2 & 12.7 \\
TEL~\cite{liang2022tree} & 79.8 & 83.7 & 51.4 & 75.7 & 67.1 & 10.7 & 72.2 & 72.9 & 43.9 & 59.7 & 50.5 & 15.2 & 76.4 & 80.1 & 48.2 & 70.9 & 62.2 & 13.3 & 71.2 & 73.5 & 42.0 & 59.5 & 50.3 & 14.0 \\
SCOD~\cite{he2023weakly} & 80.2 & 84.5 & 51.0 & 75.4 & 66.5 & 9.1 & 74.2 & 75.8 & 45.1 & 61.7 & 52.0 & 12.6 & 75.8 & 80.6 & 45.7 & 70.1 & 60.6 & 12.3 & 73.4 & 77.0 & 42.8 & 62.9 & 52.9 & 11.3 \\
SAM-WS~\cite{chen2023sam} & 65.9 & 64.9 & 26.5 & 51.6 & 40.2 & 17.5 & 60.3 & 54.6 & 23.4 & 38.8 & 28.9 & 23.7 & 60.3 & 58.0 & 20.5 & 42.3 & 31.9 & 22.0 & 56.9 & 52.3 & 20.2 & 35.2 & 25.8 & 25.1 \\
GenSAM~\cite{hu2024relax} & 77.8 & 82.9 & 47.2 & 74.2 & 64.9 & 10.6 & 72.7 & 74.3 & 42.2 & 60.9 & 51.0 & 14.3 & 73.7 & 79.0 & 42.6 & 69.2 & 59.6 & 13.6 & 71.3 & 75.0 & 39.4 & 61.3 & 51.0 & 13.3 \\
ProMaC~\cite{hu2024leveraging} & 84.7 & \underline{88.3} & 61.9 & \underline{80.9} & \underline{72.8} & 7.8 & 74.9 & 77.4 & 51.3 & 64.8 & 55.8 & 12.8 & 79.5 & 82.8 & 54.9 & \underline{73.6} & \underline{65.1} & 11.5 & 75.2 & 78.5 & 50.8 & 65.0 & 56.1 & 11.3 \\
WS-SAM~\cite{he2023weakly_sam} & 84.1 & 87.4 & 59.3 & 79.5 & 70.9 & \underline{7.6} & 77.1 & 78.5 & 50.6 & 64.4 & 55.2 & \underline{9.2} & 79.6 & \underline{83.1} & 53.1 & \underline{73.6} & 64.7 & \underline{10.9} & 76.3 & 79.7 & 49.5 & 65.7 & 56.3 & \underline{9.1} \\
WeakPolyp~\cite{zhao2025weakpolyp} & 81.2 & 85.0 & 53.8 & 76.3 & 67.5 & 9.0 & 75.1 & 76.8 & 47.3 & 63.2 & 53.6 & 12.6 & 76.4 & 80.5 & 47.9 & 70.6 & 61.2 & 12.4 & 74.5 & 78.2 & 45.2 & 64.3 & 54.4 & 11.0 \\
SEE~\cite{he2025segment} & \underline{85.2} & 87.3 & \underline{62.6} & 79.6 & 70.8 & 8.0 & \underline{77.8} & \underline{78.8} & \underline{53.3} & \underline{65.8} & \underline{56.2} & 10.2 & \underline{80.5} & 82.4 & \underline{56.6} & \underline{73.6} & 64.5 & 11.7 & \underline{78.0} & \underline{80.2} & \underline{53.0} & \underline{67.0} & \underline{57.0} & 9.2 \\
\rowcolor{c1!30}
\cellcolor{c1!30}\makecell[l]{ARTEMIS\\~~(\textbf{Ours})} & \textbf{89.7} & \textbf{92.4} & \textbf{84.1} & \textbf{85.2} & \textbf{78.3} & \textbf{3.4} & \textbf{78.6} & \textbf{79.7} & \textbf{65.9} & \textbf{66.7} & \textbf{58.7} & \textbf{4.3} & \textbf{84.5} & \textbf{87.8} & \textbf{77.3} & \textbf{78.8} & \textbf{71.3} & \textbf{7.4} & \textbf{79.6} & \textbf{81.6} & \textbf{67.5} & \textbf{68.7} & \textbf{60.9} & \textbf{4.8} \\
\midrule

\multicolumn{25}{l}{\textit{Point Supervision}} \\
\midrule
SCWS~\cite{yu2021structure} & 74.6 & 80.9 & 41.5 & 71.6 & 62.1 & 12.0 & 70.3 & 71.2 & 37.7 & 57.5 & 47.4 & 15.7 & 70.8 & 77.1 & 37.0 & 66.3 & 56.3 & 14.8 & 68.6 & 71.6 & 35.2 & 57.7 & 47.3 & 14.9 \\
TEL~\cite{liang2022tree} & 77.5 & 82.9 & 46.7 & 74.7 & 66.0 & 11.4 & 72.9 & 74.3 & 42.2 & 61.3 & 51.7 & 14.3 & 72.9 & 78.8 & 41.6 & 69.3 & 60.1 & 14.5 & 71.5 & 75.1 & 39.8 & 61.7 & 51.8 & 13.3 \\
SCOD~\cite{he2023weakly} & 75.3 & 81.6 & 43.5 & 73.2 & 64.2 & 12.4 & 71.7 & 73.2 & 40.5 & 60.2 & 50.5 & 15.4 & 71.2 & 77.5 & 39.0 & 67.8 & 58.3 & 15.5 & 70.1 & 73.8 & 37.9 & 60.2 & 50.3 & 14.6 \\
SAM-WS~\cite{chen2023sam} & 69.2 & 75.9 & 33.9 & 67.4 & 57.5 & 16.3 & 66.8 & 68.9 & 32.3 & 56.7 & 46.5 & 19.8 & 65.6 & 72.5 & 29.9 & 62.5 & 52.7 & 19.4 & 64.8 & 68.6 & 29.8 & 56.1 & 45.6 & 19.4 \\
GenSAM~\cite{hu2024relax} & 76.5 & 82.1 & 45.7 & 73.1 & 63.8 & 11.3 & 72.7 & 75.4 & 42.4 & 61.9 & 51.9 & 14.3 & 72.4 & 77.9 & 41.2 & 67.9 & 58.1 & 14.5 & 71.4 & 76.0 & 39.7 & 62.4 & 51.9 & 13.1 \\
ProMaC~\cite{hu2024leveraging} & 82.2 & 86.0 & 56.2 & 78.0 & 69.3 & 9.1 & \underline{75.2} & \underline{76.1} & \underline{49.4} & \underline{62.6} & 52.3 & 12.7 & 78.9 & 82.1 & \underline{52.6} & \underline{73.2} & \underline{64.0} & 11.8 & \underline{75.2} & \underline{77.6} & \underline{49.3} & 63.8 & 54.1 & \underline{11.0} \\
WS-SAM~\cite{he2023weakly_sam} & 81.0 & 85.4 & 52.5 & 77.6 & 69.0 & 9.4 & 74.5 & 76.0 & 45.4 & 62.0 & \underline{52.6} & \underline{12.0} & 75.9 & 80.4 & 46.8 & 71.1 & 62.3 & 13.1 & 73.7 & 77.3 & 43.9 & \underline{63.9} & \underline{54.5} & 11.6 \\
SEE~\cite{he2025segment} & \underline{83.7} & \underline{87.3} & \underline{57.2} & \underline{79.4} & \underline{70.8} & \underline{8.0} & 74.5 & 75.6 & 47.1 & 62.1 & \underline{52.6} & 12.4 & \underline{79.1} & \underline{82.6} & 50.9 & 73.1 & 63.9 & \underline{11.2} & 74.2 & 77.0 & 46.1 & 63.1 & 53.6 & 11.2 \\
\rowcolor{c1!30}
\cellcolor{c1!30}\makecell[l]{ARTEMIS\\~~(\textbf{Ours})} & \textbf{86.3} & \textbf{88.7} & \textbf{65.4} & \textbf{81.2} & \textbf{73.2} & \textbf{6.8} & \textbf{77.0} & \textbf{76.9} & \textbf{52.9} & \textbf{62.8} & \textbf{53.9} & \textbf{8.0} & \textbf{81.5} & \textbf{84.0} & \textbf{59.2} & \textbf{74.8} & \textbf{66.0} & \textbf{10.0} & \textbf{77.0} & \textbf{78.6} & \textbf{52.7} & \textbf{64.4} & \textbf{55.5} & \textbf{8.0} \\
\bottomrule
\end{tabular}
}
\vspace{-4mm}
\label{tab:SUN-SEG-Combined-Weakly}
\end{table*}

\begin{table}[h]
\centering
\caption{Quantitative comparison on the \textbf{CVC-ClinicDB-612} dataset under \textit{Fully Supervised}, \textit{Scribble Supervision}, and \textit{Point Supervision} settings. The best and second-best results are highlighted in \textbf{bold} and \underline{underlined}, respectively.}
\vspace{-2mm}
\setlength{\tabcolsep}{6pt}
\renewcommand{\arraystretch}{0.80}
\resizebox{\columnwidth}{!}{
\begin{tabular}{l|cccccc}
\toprule
\multirow{2}{*}{Method} & \multicolumn{6}{c}{CVC-ClinicDB-612 (\%)} \\
& $S_\alpha\uparrow$ & $E_{\phi}^{mn}\uparrow$ & $F_\beta^w\uparrow$ & Dice$\uparrow$ & IoU$\uparrow$ & MAE$\downarrow$ \\
\midrule

\multicolumn{7}{l}{\textit{Fully Supervised}} \\
\midrule
PraNet~\cite{fan2020pranet} & 93.6 & 95.7 & 88.6 & 92.3 & 86.9 & 0.8 \\
PNS~\cite{ji2021progressively} & 93.0 & 95.3 & 87.2 & 92.0 & 86.1 & 0.8 \\
PNS+~\cite{ji2022video} & 94.5 & 96.7 & 89.5 & 93.1 & 88.3 & 0.9 \\
\midrule

\multicolumn{7}{l}{\textit{Scribble Supervision}} \\
\midrule
SCWS~\cite{yu2021structure} & 83.1 & 84.4 & 68.1 & 71.2 & 58.1 & 2.2 \\
TEL~\cite{liang2022tree} & 83.1 & 84.1 & 67.8 & 70.5 & 55.0 & 2.1 \\
SCOD~\cite{he2023weakly} & 83.3 & 85.5 & 69.2 & 73.5 & 59.1 & 2.1 \\
SAM-WS~\cite{chen2023sam} & 72.9 & 73.1 & 53.1 & 52.4 & 37.6 & 6.7 \\
GenSAM~\cite{hu2024relax} & 80.1 & 82.2 & 67.1 & 68.1 & 52.2 & 2.9 \\
ProMaC~\cite{hu2024leveraging} & 88.2 & 91.4 & \underline{80.2} & \underline{82.9} & \underline{71.1} & 1.8 \\
WS-SAM~\cite{he2023weakly_sam} & 87.6 & 89.3 & 76.8 & 79.1 & 67.6 & 1.8 \\
WeakPolyp~\cite{zhao2025weakpolyp} & 83.5 & 85.5 & 69.8 & 72.6 & 59.9 & 1.9 \\
SEE~\cite{he2025segment} & \underline{89.8} & \underline{92.1} & 78.9 & 81.9 & 70.4 & \underline{1.6} \\
\rowcolor{c1!30}
\cellcolor{c1!30}\makecell[l]{ARTEMIS\\~~(\textbf{Ours})} & \textbf{91.7} & \textbf{94.8} & \textbf{86.5} & \textbf{88.7} & \textbf{80.2} & \textbf{1.1} \\
\midrule

\multicolumn{7}{l}{\textit{Point Supervision}} \\
\midrule
SCWS~\cite{yu2021structure} & 76.1 & 80.3 & 62.1 & 64.7 & 48.7 & 5.1 \\
TEL~\cite{liang2022tree} & 77.7 & 82.4 & 63.7 & 70.1 & 54.7 & 4.2 \\
SCOD~\cite{he2023weakly} & 77.1 & 81.1 & 63.5 & 69.5 & 54.2 & 4.9 \\
SAM-WS~\cite{chen2023sam} & 69.5 & 68.0 & 46.7 & 48.1 & 33.1 & 6.4 \\
GenSAM~\cite{hu2024relax} & 75.2 & 79.1 & 62.1 & 62.7 & 46.2 & 5.0 \\
ProMaC~\cite{hu2024leveraging} & 83.8 & 83.1 & 69.1 & 70.9 & 55.5 & 2.7 \\
WS-SAM~\cite{he2023weakly_sam} & 80.5 & 84.1 & 66.7 & 70.2 & 55.7 & 3.2 \\
SEE~\cite{he2025segment} & \underline{84.1} & \underline{85.2} & \underline{73.2} & \underline{75.1} & \underline{61.4} & \underline{2.0} \\
\rowcolor{c1!30}
\cellcolor{c1!30}\makecell[l]{ARTEMIS\\~~(\textbf{Ours})} & \textbf{88.7} & \textbf{90.7} & \textbf{79.4} & \textbf{83.1} & \textbf{72.3} & \textbf{1.7} \\
\bottomrule
\end{tabular}
}
\vspace{-6mm}
\label{tab:CVC-Combined-Single-Weakly}
\end{table}

\begin{table*}[t]
\centering
\caption{Quantitative comparison on the \textbf{SUN-SEG-Easy} and \textbf{SUN-SEG-Hard} datasets under \textit{1/8 Labeled Training Data} and \textit{1/16 Labeled Training Data} semi-supervised settings. The best and second-best results are highlighted in \textbf{bold} and \underline{underlined}, respectively.}
\vspace{-2mm}
\setlength{\tabcolsep}{0.005pt}
\renewcommand{\arraystretch}{0.93}
\resizebox{\textwidth}{!}{
\begin{tabular}{l|cccccc|cccccc|cccccc|cccccc}
\toprule
\multirow{2}{*}{Method} & \multicolumn{6}{c|}{SUN-SEG-Easy-Seen (\%)} & \multicolumn{6}{c|}{SUN-SEG-Easy-Unseen (\%)} & \multicolumn{6}{c|}{SUN-SEG-Hard-Seen (\%)} & \multicolumn{6}{c}{SUN-SEG-Hard-Unseen (\%)} \\
& $S_\alpha\uparrow$ & $E_{\phi}^{mn}\uparrow$ & $F_\beta^w\uparrow$ & Dice$\uparrow$ & IoU$\uparrow$ & MAE$\downarrow$ & $S_\alpha\uparrow$ & $E_{\phi}^{mn}\uparrow$ & $F_\beta^w\uparrow$ & Dice$\uparrow$ & IoU$\uparrow$ & MAE$\downarrow$ & $S_\alpha\uparrow$ & $E_{\phi}^{mn}\uparrow$ & $F_\beta^w\uparrow$ & Dice$\uparrow$ & IoU$\uparrow$ & MAE$\downarrow$ & $S_\alpha\uparrow$ & $E_{\phi}^{mn}\uparrow$ & $F_\beta^w\uparrow$ & Dice$\uparrow$ & IoU$\uparrow$ & MAE$\downarrow$ \\
\midrule

\multicolumn{25}{l}{\textit{1/8 Labeled Training Data}} \\
\midrule
MT~\cite{tarvainen2017mean} & 75.2 & 80.5 & 44.7 & 72.5 & 63.3 & 13.4 & 71.7 & 74.1 & 42.3 & 63.0 & 53.5 & 17.4 & 71.0 & 76.0 & 39.9 & 67.1 & 57.5 & 16.5 & 70.7 & 74.8 & 40.0 & 63.7 & 53.5 & 16.0 \\
SAM-S~\cite{chen2023sam} & 67.6 & 71.0 & 28.7 & 61.0 & 50.1 & 17.4 & 63.1 & 60.5 & 26.3 & 48.0 & 37.5 & 22.9 & 63.5 & 66.7 & 23.4 & 55.0 & 43.8 & 19.8 & 61.7 & 60.2 & 24.4 & 46.9 & 36.3 & 22.0 \\
MCF~\cite{wang2023mcf} & 80.2 & 85.2 & 53.1 & 76.9 & 68.1 & 9.8 & 75.6 & 77.5 & 49.2 & 65.0 & 55.6 & 13.1 & 77.4 & 81.9 & 50.0 & 72.5 & 63.1 & 12.2 & 75.6 & 79.0 & 48.1 & 66.5 & 56.6 & 11.4 \\
CauSSL~\cite{miao2023caussl} & 83.8 & 87.3 & 57.2 & 79.7 & 71.4 & 8.4 & 75.6 & 76.3 & 47.8 & 62.4 & 53.1 & 11.2 & 78.8 & 81.9 & 51.2 & 72.9 & 64.2 & 12.0 & 76.2 & 78.3 & 47.9 & 64.9 & 55.6 & 10.6 \\
CML~\cite{wu2024cross} & 85.5 & 87.4 & 63.0 & 79.7 & 71.2 & 7.9 & \underline{78.5} & \underline{79.6} & 51.4 & 65.6 & 56.2 & 8.1 & 80.3 & 83.1 & 55.8 & 73.7 & 64.8 & 11.2 & \underline{79.3} & \underline{81.2} & 53.5 & \underline{67.6} & 57.9 & 7.4 \\
AD-MT~\cite{zhao2024alternate} & 85.1 & 88.0 & 61.8 & 79.8 & 71.0 & 7.4 & 77.9 & 79.1 & 52.8 & 65.7 & 56.2 & 9.9 & 80.6 & 83.2 & 55.8 & 74.0 & 64.9 & 10.8 & 77.6 & 80.0 & 51.7 & 66.2 & 56.5 & 9.1 \\
ST-SAM~\cite{hu2025st} & 86.7 & 89.1 & 67.3 & 81.4 & 73.5 & 6.7 & 76.7 & 76.8 & 52.8 & 61.8 & 53.2 & 7.2 & 81.4 & 83.8 & 60.1 & 74.3 & 66.0 & 10.1 & 77.4 & 79.0 & 54.1 & 64.0 & 55.5 & 7.2 \\
KnowSAM~\cite{huang2025learnable} & 87.0 & 89.8 & 80.3 & 81.6 & 74.1 & 5.3 & 75.5 & 74.9 & 60.4 & 60.9 & 52.9 & \underline{5.0} & 81.7 & 83.8 & 73.3 & 74.6 & 66.7 & 9.0 & 74.9 & 74.5 & 59.6 & 60.2 & 52.3 & 5.7 \\
SEE~\cite{he2025segment} & \underline{88.1} & \underline{91.3} & \underline{81.9} & \underline{83.9} & \underline{76.7} & \underline{4.9} & 77.9 & \underline{79.6} & \underline{65.0} & \underline{66.1} & \underline{58.2} & 5.4 & \underline{82.9} & \underline{86.4} & \underline{74.6} & \underline{77.1} & \underline{68.9} & \underline{8.7} & 78.7 & 81.0 & \underline{65.8} & 67.2 & \underline{59.3} & \underline{5.5} \\
\rowcolor{c1!30}
\cellcolor{c1!30}\makecell[l]{ARTEMIS\\~~(\textbf{Ours})} & \textbf{90.7} & \textbf{93.6} & \textbf{85.2} & \textbf{86.5} & \textbf{79.9} & \textbf{2.7} & \textbf{79.4} & \textbf{81.1} & \textbf{67.1} & \textbf{68.2} & \textbf{60.4} & \textbf{4.8} & \textbf{86.2} & \textbf{90.1} & \textbf{78.2} & \textbf{80.0} & \textbf{72.2} & \textbf{4.5} & \textbf{80.8} & \textbf{83.9} & \textbf{69.4} & \textbf{70.9} & \textbf{63.1} & \textbf{4.6} \\
\midrule

\multicolumn{25}{l}{\textit{1/16 Labeled Training Data}} \\
\midrule
MT~\cite{tarvainen2017mean} & 74.4 & 79.9 & 41.4 & 70.4 & 60.6 & 12.3 & 70.2 & 71.6 & 37.7 & 58.1 & 48.0 & 16.1 & 71.6 & 77.0 & 37.6 & 66.4 & 56.5 & 14.5 & 68.5 & 71.7 & 34.8 & 57.9 & 47.4 & 15.3 \\
SAM-S~\cite{chen2023sam} & 65.6 & 68.9 & 26.9 & 57.6 & 46.6 & 18.5 & 61.5 & 60.3 & 24.5 & 45.1 & 34.6 & 22.6 & 61.1 & 64.7 & 22.1 & 50.9 & 40.1 & 21.2 & 58.5 & 58.0 & 21.4 & 41.8 & 31.6 & 22.9 \\
MCF~\cite{wang2023mcf} & 77.7 & 83.1 & 48.3 & 75.1 & 66.3 & 11.3 & 72.0 & 73.5 & 43.5 & 62.2 & 52.7 & 16.5 & 73.3 & 78.9 & 43.0 & 69.3 & 60.3 & 14.2 & 70.5 & 74.4 & 40.4 & 61.6 & 51.9 & 14.8 \\
CauSSL~\cite{miao2023caussl} & 81.7 & 86.3 & 54.1 & 78.4 & 69.6 & 8.7 & 75.4 & 77.8 & 46.7 & 63.8 & 54.6 & 11.1 & 77.2 & 81.9 & 48.3 & 72.7 & 63.7 & 11.9 & 74.5 & 78.7 & 45.5 & 65.1 & 55.6 & 10.6 \\
CML~\cite{wu2024cross} & 83.7 & 87.8 & 58.6 & 80.1 & 71.9 & 7.7 & 75.6 & 77.1 & 48.6 & 63.0 & 53.9 & 10.4 & 78.7 & 82.9 & 51.7 & 73.6 & 64.8 & 11.0 & 74.7 & 78.2 & 47.3 & 64.1 & 54.9 & 10.0 \\
AD-MT~\cite{zhao2024alternate} & 83.9 & 87.9 & 58.8 & 80.1 & 71.8 & 7.7 & 76.3 & 78.2 & 50.0 & \underline{64.5} & \underline{55.3} & 10.6 & 79.1 & 83.0 & 52.1 & 73.8 & 64.9 & 11.0 & 75.8 & 79.3 & 49.0 & 65.7 & 56.3 & 9.9 \\
ST-SAM~\cite{hu2025st} & 84.5 & 87.9 & 60.1 & 80.1 & 71.9 & 7.5 & 76.6 & 77.5 & 50.1 & 63.0 & 54.0 & 8.8 & 79.9 & 83.3 & 53.9 & 74.0 & 65.2 & 10.7 & 76.2 & 78.9 & 49.3 & 64.6 & 55.5 & 8.7 \\
KnowSAM~\cite{huang2025learnable} & 85.0 & \underline{88.5} & 62.0 & \underline{80.9} & \underline{72.8} & 7.1 & \underline{77.6} & \underline{78.5} & 52.2 & 64.0 & 55.2 & 7.6 & 80.5 & \underline{84.2} & 56.2 & \underline{74.8} & \underline{66.1} & 10.3 & 77.2 & \underline{80.3} & 51.6 & \underline{65.8} & \underline{56.8} & 7.5 \\
SEE~\cite{he2025segment} & \underline{86.1} & 87.4 & \underline{68.2} & 78.5 & 69.9 & \underline{6.6} & 76.6 & 76.4 & \underline{55.0} & 61.5 & 52.6 & \underline{7.0} & \underline{82.0} & 83.7 & \underline{62.8} & 73.4 & 64.7 & \underline{9.8} & \underline{77.5} & 78.3 & \underline{55.5} & 63.1 & 53.9 & \underline{6.7} \\
\rowcolor{c1!30}
\cellcolor{c1!30}\makecell[l]{ARTEMIS\\~~(\textbf{Ours})} & \textbf{89.9} & \textbf{92.4} & \textbf{84.3} & \textbf{85.5} & \textbf{78.9} & \textbf{3.4} & \textbf{78.3} & \textbf{79.8} & \textbf{66.0} & \textbf{66.9} & \textbf{58.9} & \textbf{4.9} & \textbf{84.5} & \textbf{87.9} & \textbf{77.4} & \textbf{79.1} & \textbf{71.5} & \textbf{7.6} & \textbf{79.2} & \textbf{81.4} & \textbf{67.1} & \textbf{68.2} & \textbf{60.2} & \textbf{5.0} \\
\bottomrule
\end{tabular}
}
\vspace{-4mm}
\label{tab:SUN-SEG-Combined-Semi}
\end{table*}

\begin{table}[h]
\centering
\caption{Quantitative comparison on the \textbf{CVC-ClinicDB-612} dataset under \textit{1/8 Labeled Training Data} and \textit{1/16 Labeled Training Data} semi-supervised settings. The best and second-best results are highlighted in \textbf{bold} and \underline{underlined}, respectively.}
\vspace{-2mm}
\setlength{\tabcolsep}{6pt}
\renewcommand{\arraystretch}{0.80}
\resizebox{\columnwidth}{!}{
\begin{tabular}{l|cccccc}
\toprule
\multirow{2}{*}{Method} & \multicolumn{6}{c}{CVC-ClinicDB-612 (\%)} \\
& $S_\alpha\uparrow$ & $E_{\phi}^{mn}\uparrow$ & $F_\beta^w\uparrow$ & Dice$\uparrow$ & IoU$\uparrow$ & MAE$\downarrow$ \\
\midrule

\multicolumn{7}{l}{\textit{1/8 Labeled Training Data}} \\
\midrule
MT~\cite{tarvainen2017mean} & 80.2 & 84.2 & 65.7 & 69.1 & 52.7 & 4.3 \\
SAM-S~\cite{chen2023sam} & 70.1 & 69.2 & 48.7 & 49.7 & 33.5 & 5.7 \\
MCF~\cite{wang2023mcf} & 84.2 & 85.3 & 73.1 & 75.4 & 60.5 & 1.9 \\
CauSSL~\cite{miao2023caussl} & 85.9 & 86.7 & 74.5 & 77.5 & 63.8 & 1.7 \\
CML~\cite{wu2024cross} & 88.1 & 92.1 & 78.8 & 80.7 & 67.3 & 1.7 \\
AD-MT~\cite{zhao2024alternate} & 87.5 & 90.1 & 79.1 & 81.5 & 68.9 & 1.9 \\
ST-SAM~\cite{hu2025st} & 88.5 & 92.4 & 80.5 & 82.8 & 70.7 & 1.8 \\
KnowSAM~\cite{huang2025learnable} & 89.5 & 92.8 & 81.7 & 84.1 & 72.1 & 1.6 \\
SEE~\cite{he2025segment} & \underline{90.1} & \underline{93.1} & \underline{84.7} & \underline{86.7} & \underline{76.1} & \underline{1.3} \\
\rowcolor{c1!30}
\cellcolor{c1!30}\makecell[l]{ARTEMIS\\~~(\textbf{Ours})} & \textbf{92.8} & \textbf{95.2} & \textbf{88.2} & \textbf{89.6} & \textbf{81.7} & \textbf{0.8} \\
\midrule

\multicolumn{7}{l}{\textit{1/16 Labeled Training Data}} \\
\midrule
MT~\cite{tarvainen2017mean} & 78.1 & 82.8 & 63.7 & 65.1 & 49.2 & 4.6 \\
SAM-S~\cite{chen2023sam} & 68.5 & 69.0 & 48.1 & 49.1 & 32.1 & 5.9 \\
MCF~\cite{wang2023mcf} & 82.4 & 83.3 & 66.5 & 69.1 & 52.5 & 2.3 \\
CauSSL~\cite{miao2023caussl} & 84.1 & 84.7 & 72.1 & 74.9 & 58.1 & 1.9 \\
CML~\cite{wu2024cross} & 86.9 & 88.1 & 75.5 & 78.2 & 64.5 & 1.8 \\
AD-MT~\cite{zhao2024alternate} & 86.2 & 88.2 & 76.1 & 77.6 & 63.2 & 1.9 \\
ST-SAM~\cite{hu2025st} & 87.5 & 91.3 & 79.1 & 79.3 & 65.5 & 1.8 \\
KnowSAM~\cite{huang2025learnable} & 88.1 & \underline{92.5} & \underline{81.5} & 80.1 & 66.9 & 1.8 \\
SEE~\cite{he2025segment} & \underline{88.5} & 92.3 & 78.3 & \underline{81.3} & \underline{67.1} & \underline{1.5} \\
\rowcolor{c1!30}
\cellcolor{c1!30}\makecell[l]{ARTEMIS\\~~(\textbf{Ours})} & \textbf{90.9} & \textbf{92.9} & \textbf{85.2} & \textbf{84.8} & \textbf{76.1} & \textbf{1.2} \\
\bottomrule
\end{tabular}
}
\vspace{-6mm}
\label{tab:CVC-Combined-Single-Semi}
\end{table}

\subsection{Experimental Settings}
\minisection{Datasets and supervision settings.}
Experiments are conducted on SUN-SEG~\cite{ji2022video} and CVC-ClinicDB-612~\cite{bernal2015wm}. For SUN-SEG, we follow the standard Easy/Hard and Seen/Unseen protocol, yielding four splits: SUN-SEG-Easy-Seen, SUN-SEG-Easy-Unseen, SUN-SEG-Hard-Seen, and SUN-SEG-Hard-Unseen. CVC-ClinicDB-612 serves as an additional benchmark and is split into training/testing sets at a 6:4 ratio.
We evaluate ARTEMIS under two imperfectly supervised regimes: \textbf{1) weak supervision}, including scribble and point annotations; and \textbf{2) semi-supervision}, using only 1/8 or 1/16 densely labeled SUN-SEG training data following common practice.

\minisection{Evaluation metrics.}
We report six widely used metrics: structure measure $S_\alpha$~\cite{fan2017structure}, mean E-measure $E_{\phi}^{mn}$~\cite{fan2018enhanced}, weighted F-measure $F_\beta^w$~\cite{margolin2014evaluate}, Dice, IoU, and mean absolute error (MAE). Larger values indicate better performance for $S_\alpha$, $E_{\phi}^{mn}$, $F_\beta^w$, Dice, and IoU, while smaller values are better for MAE.

\minisection{Compared methods.}
For weak supervision, we compare with SCWS~\cite{yu2021structure}, TEL~\cite{liang2022tree}, SCOD~\cite{he2023weakly}, SAM-WS~\cite{chen2023sam}, GenSAM~\cite{hu2024relax}, ProMaC~\cite{hu2024leveraging}, WS-SAM~\cite{he2023weakly_sam}, WeakPolyp~\cite{zhao2025weakpolyp}, and SEE~\cite{he2025segment}. For semi-supervision, we compare with MT~\cite{tarvainen2017mean}, SAM-S~\cite{chen2023sam}, MCF~\cite{wang2023mcf}, CauSSL~\cite{miao2023caussl}, CML~\cite{wu2024cross}, AD-MT~\cite{zhao2024alternate}, ST-SAM~\cite{hu2025st}, KnowSAM~\cite{huang2025learnable}, and SEE~\cite{he2025segment}. PraNet~\cite{fan2020pranet}, PNS~\cite{ji2021progressively}, and PNS+~\cite{ji2022video} are fully supervised methods. We trained these methods under the same settings.

\subsection{Comparison to State-of-the-Arts}
\minisection{Weakly supervised comparison.}
\cref{tab:SUN-SEG-Combined-Weakly,tab:CVC-Combined-Single-Weakly} report the scribble and point results. ARTEMIS ranks first among weakly supervised methods on both datasets. Three observations are noteworthy. \textbf{1)} On SUN-SEG with scribble supervision, ARTEMIS improves Dice from 80.9\% to 85.2\% on Easy-Seen and from 67.0\% to 68.7\% on Hard-Unseen over the strongest competitor. \textbf{2)} The $F_\beta^w$ gains are larger, from 62.6\% to 84.1\% and from 53.0\% to 67.5\%, indicating more complete structures. ARTEMIS also reduces MAE from 7.6\% to 3.4\% on Easy-Seen and from 9.1\% to 4.8\% on Hard-Unseen, suggesting fewer background leaks and missed regions. \textbf{3)} On CVC-ClinicDB-612, ARTEMIS raises Dice from 82.9\% to 88.7\% under scribble supervision and from 75.1\% to 83.1\% under point supervision, while IoU increases to 80.2\% and 72.3\%, respectively.
These gains support our motivation: agent-guided mask evolution turns sparse annotations into reliable temporal anchors and dense evolved pseudo masks.

\minisection{Semi-supervised comparison.}
\cref{tab:SUN-SEG-Combined-Semi,tab:CVC-Combined-Single-Semi} show the 1/8 and 1/16 labeled-data results. ARTEMIS achieves the best performance in all semi-supervised settings. \textbf{1)} On SUN-SEG-Easy-Seen, Dice increases from 83.9\% to 86.5\% with 1/8 labels and from 80.9\% to 85.5\% with 1/16 labels. \textbf{2)} On the harder SUN-SEG-Hard-Unseen split, Dice improves from 67.6\% to 70.9\% and from 65.8\% to 68.2\%, respectively; the corresponding $F_\beta^w$ scores also rise to 69.4\% and 67.1\%, showing stronger structure recovery under scarce labels. \textbf{3)} On CVC-ClinicDB-612, ARTEMIS improves Dice from 86.7\% to 89.6\% and from 81.3\% to 84.8\%, with IoU gains from 76.1\% to 81.7\% and from 67.1\% to 76.1\%. This confirms that anchor-based bidirectional evolution can densify missing frame-level labels into temporally consistent masks for semi-supervised learning. The consistent gains over teacher-student and SAM-based baselines further indicate that reliability-aware training is important after pseudo-label completion, because evolved masks still contain noise.

\begin{figure*}[!t]
    \centering
    \includegraphics[width=0.95\linewidth]{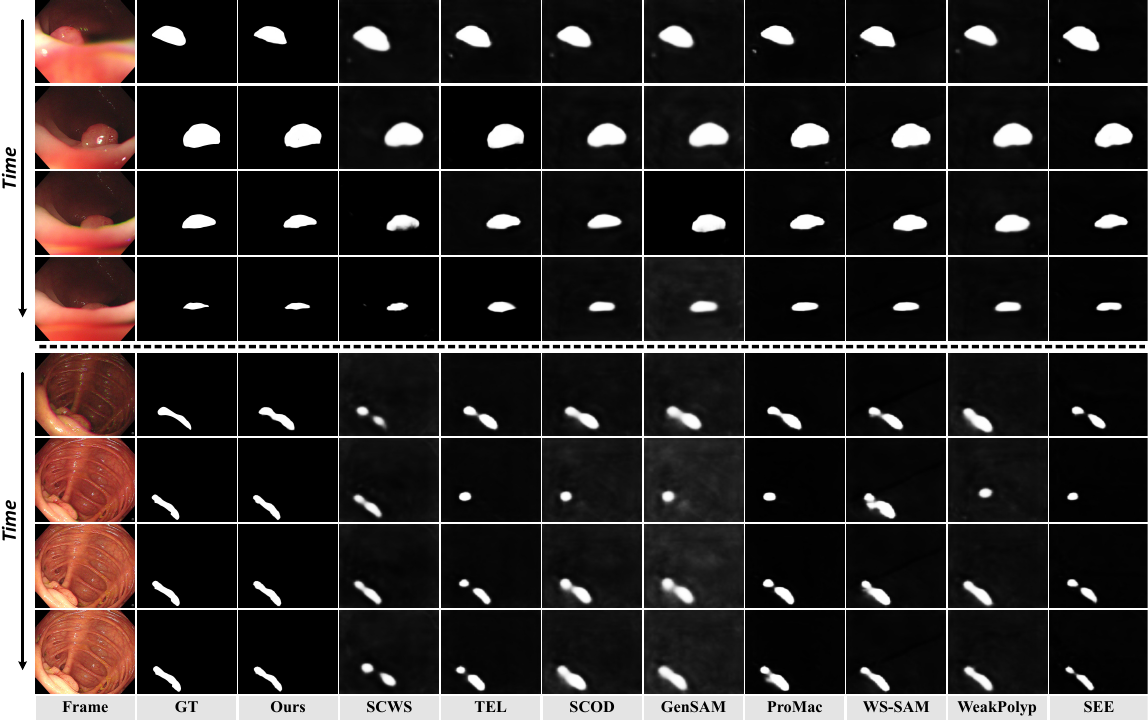}
    \vspace{-3mm}
    \caption{\textbf{Qualitative comparison.} Under \textit{Scribble Supervision}, rows 1--4 show blurry polyps with temporal drift, where ARTEMIS better preserves accurate boundaries. Rows 5--8 show a small polyp target, where ARTEMIS produces more continuous and precise predictions across frames.}
    \label{fig:qualitative_compairson_scribble}

    \vspace{1mm}

    \includegraphics[width=0.95\linewidth]{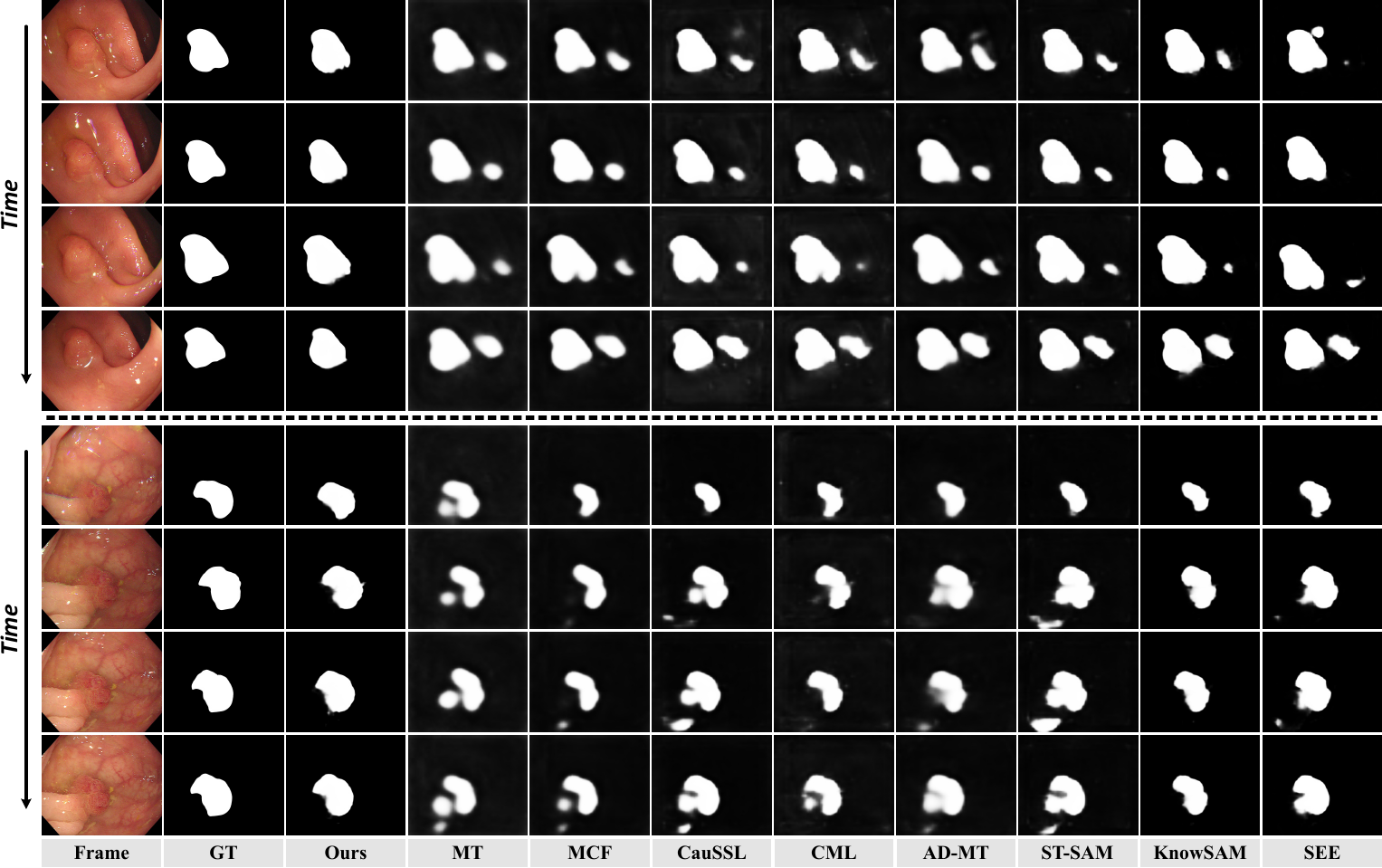}
    \vspace{-3mm}
    \caption{\textbf{Qualitative comparison.} Under the \textit{1/8 Labeled Training Data} setting, the two video clips, each with four frames, contain polyps highly similar to the background. ARTEMIS avoids the over-segmentation and under-segmentation observed in competing methods and identifies the polyp regions accurately.}
    \label{fig:qualitative_compairson_semi}
\end{figure*}

\begin{figure*}[!t]
    \centering
    \includegraphics[width=\linewidth]{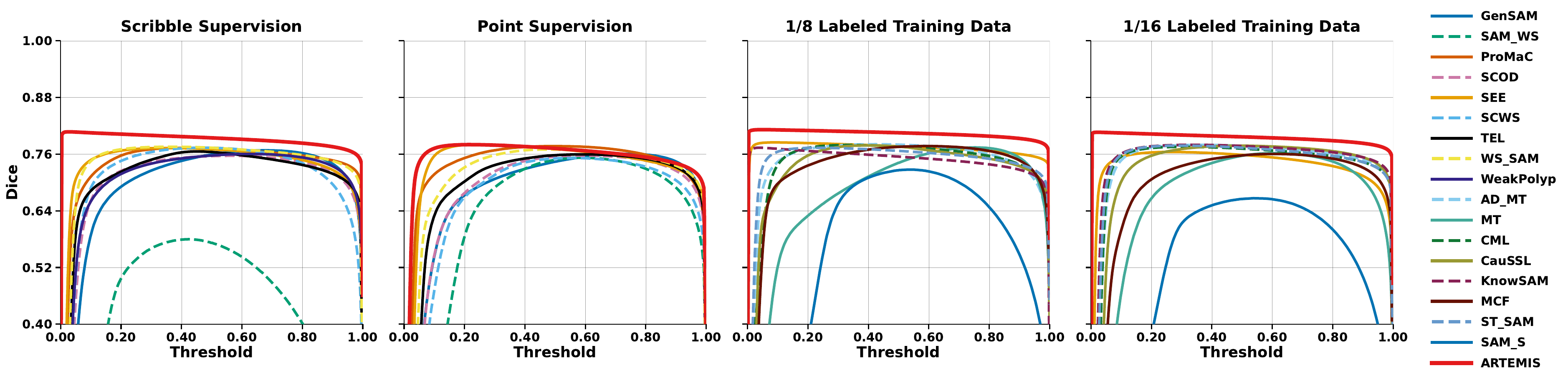}
    \vspace{-9mm}
    \caption{\textbf{Comparison of Dice-threshold curves} on the \textbf{SUN-SEG-Hard-Seen} across four supervision settings. From left to right, the subplots show results under \textit{Scribble Supervision}, \textit{Point Supervision}, \textit{1/8 Labeled Training Data}, and \textit{1/16 Labeled Training Data}.}
    \label{fig:compare_testhard_seen_dice_1x4}
    \vspace{0.5mm}

    \includegraphics[width=\linewidth]{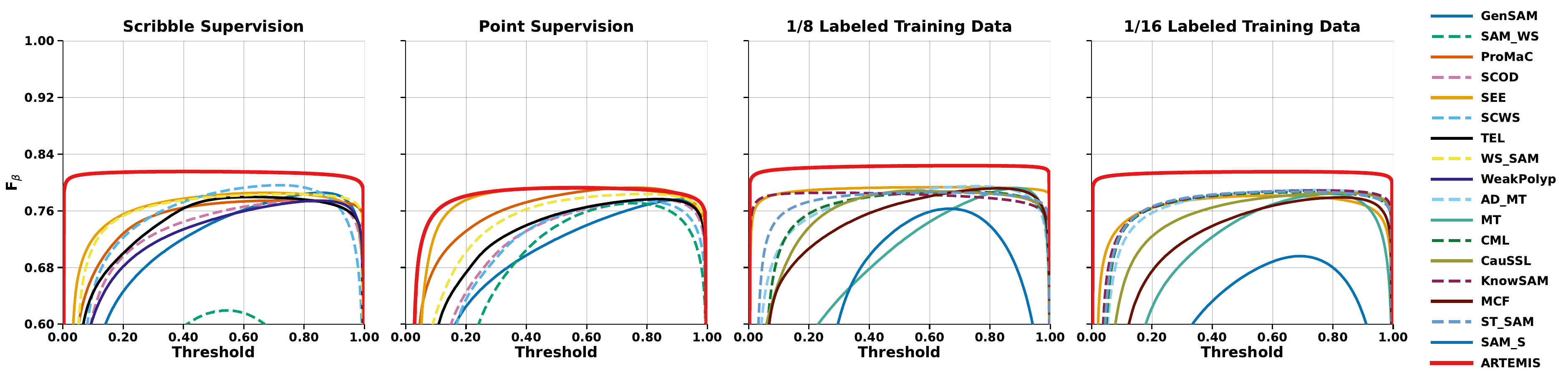}
    \vspace{-9mm}
    \caption{\textbf{Comparison of $F_\beta$-threshold curves} on the \textbf{SUN-SEG-Hard-Seen} across four supervision settings. From left to right, the subplots show results under \textit{Scribble Supervision}, \textit{Point Supervision}, \textit{1/8 Labeled Training Data}, and \textit{1/16 Labeled Training Data}.}
    \label{fig:compare_testhard_seen_fm_1x4}
    \vspace{-6mm}
\end{figure*}

\minisection{Threshold curves and qualitative comparison.}
\cref{fig:compare_testhard_seen_dice_1x4,fig:compare_testhard_seen_fm_1x4} compare Dice- and $F_\beta$-threshold curves on SUN-SEG-Hard-Seen. ARTEMIS remains stronger across thresholds under all four settings, suggesting that its gains are not tied to a specific binarization threshold. Qualitatively, \cref{fig:qualitative_compairson_scribble} shows that ARTEMIS better preserves blurry, drifting, and small polyp regions, while competing methods often produce inaccurate or temporally discontinuous masks. \cref{fig:qualitative_compairson_semi} further shows that ARTEMIS reduces over- and under-segmentation for background-like polyps by combining reliable pseudo supervision with reference-guided temporal identity modeling.

\subsection{Ablation Studies}

We ablate ARTEMIS following its two-stage pipeline. \textbf{Stage 1} evaluates reliable anchor selection and bidirectional mask evolution, while \textbf{Stage 2} analyzes reference selection, reliability-aware robust loss, and RPTM.

\begin{table*}[t]
\centering
\caption{Overall component analysis on \textbf{SUN-SEG} dataset under \textit{Scribble Supervision} and \textit{1/8 Labeled Training Data} settings. B: baseline segmenter; Evo: agent-guided bidirectional mask evolution; Ref: reliability-guided reference selection; Rel: reliability-aware robust loss; RPTM: Reference Prototype Transport Module.}
\label{tab:overall_components}
\vspace{-2mm}
\setlength{\tabcolsep}{3.5pt} 
\renewcommand{\arraystretch}{0.85}
\resizebox{\textwidth}{!}{
\begin{tabular}{ccccc|cccc|cccc|cccc|cccc}
\toprule
\multirow{2}{*}{B} & \multirow{2}{*}{Evo} & \multirow{2}{*}{Ref} & \multirow{2}{*}{Rel} & \multirow{2}{*}{RPTM} & \multicolumn{4}{c|}{SUN-SEG-Easy-Seen (\%)} & \multicolumn{4}{c|}{SUN-SEG-Easy-Unseen (\%)} & \multicolumn{4}{c|}{SUN-SEG-Hard-Seen (\%)} & \multicolumn{4}{c}{SUN-SEG-Hard-Unseen (\%)} \\
& & & & & $S_\alpha\uparrow$ & $E_{\phi}^{mn}\uparrow$ & $F_\beta^w\uparrow$ & Dice$\uparrow$ & $S_\alpha\uparrow$ & $E_{\phi}^{mn}\uparrow$ & $F_\beta^w\uparrow$ & Dice$\uparrow$ & $S_\alpha\uparrow$ & $E_{\phi}^{mn}\uparrow$ & $F_\beta^w\uparrow$ & Dice$\uparrow$ & $S_\alpha\uparrow$ & $E_{\phi}^{mn}\uparrow$ & $F_\beta^w\uparrow$ & Dice$\uparrow$ \\
\midrule

\multicolumn{21}{l}{\textit{Scribble Supervision}} \\
\midrule
$\checkmark$ & -- & -- & -- & -- & 67.0 & 67.9 & 28.0 & 57.0 & 62.9 & 59.3 & 26.6 & 46.0 & 62.0 & 62.0 & 22.5 & 49.0 & 60.5 & 56.9 & 23.7 & 42.9 \\
$\checkmark$ & $\checkmark$ & -- & -- & -- & 81.1 & 85.2 & 55.1 & 77.7 & 74.4 & 76.2 & 48.1 & 63.4 & 76.9 & 80.7 & 50.4 & 71.8 & 73.7 & 77.4 & 45.9 & 63.7 \\
$\checkmark$ & $\checkmark$ & $\checkmark$ & -- & -- & 84.2 & 87.3 & 61.0 & 79.7 & 77.4 & 79.1 & 53.6 & 64.3 & 80.1 & 82.9 & 56.0 & 74.5 & 78.0 & 80.8 & 53.1 & 65.0 \\
$\checkmark$ & $\checkmark$ & $\checkmark$ & $\checkmark$ & -- & 87.6 & 89.6 & 77.1 & 81.8 & 77.6 & 78.5 & 60.9 & 64.5 & 82.9 & 85.2 & 70.5 & 75.8 & 78.7 & 79.8 & 62.2 & 65.7 \\
\rowcolor{c1!30}
$\checkmark$ & $\checkmark$ & $\checkmark$ & $\checkmark$ & $\checkmark$ & 89.7 & 92.4 & 84.1 & 85.2 & 78.6 & 79.7 & 65.9 & 66.7 & 84.5 & 87.8 & 77.3 & 78.8 & 79.6 & 81.6 & 67.5 & 68.7 \\
\midrule

\multicolumn{21}{l}{\textit{1/8 Labeled Training Data}} \\
\midrule
$\checkmark$ & -- & -- & -- & -- & 69.5 & 75.5 & 34.6 & 66.0 & 67.0 & 69.0 & 32.8 & 56.3 & 66.0 & 72.0 & 30.5 & 61.6 & 65.2 & 68.5 & 30.5 & 55.3 \\
$\checkmark$ & $\checkmark$ & -- & -- & -- & 83.5 & 86.9 & 58.1 & 80.0 & 73.9 & 74.1 & 46.2 & 61.1 & 79.2 & 82.7 & 49.8 & 72.1 & 73.8 & 77.1 & 45.9 & 63.7 \\
$\checkmark$ & $\checkmark$ & $\checkmark$ & -- & -- & 86.1 & 87.9 & 65.1 & 81.1 & 77.4 & 76.5 & 52.1 & 65.5 & 81.7 & 84.1 & 60.2 & 75.1 & 76.1 & 77.2 & 51.5 & 64.1 \\
$\checkmark$ & $\checkmark$ & $\checkmark$ & $\checkmark$ & -- & 88.3 & 91.6 & 82.7 & 84.4 & 78.0 & 79.7 & 65.3 & 66.1 & 83.1 & 86.7 & 75.5 & 77.3 & 78.7 & 81.1 & 66.0 & 67.2 \\
\rowcolor{c1!30}
$\checkmark$ & $\checkmark$ & $\checkmark$ & $\checkmark$ & $\checkmark$ & 90.7 & 93.6 & 85.2 & 86.5 & 79.4 & 81.1 & 67.1 & 68.2 & 86.2 & 90.1 & 78.2 & 80.0 & 80.8 & 83.9 & 69.4 & 70.9 \\
\bottomrule
\end{tabular}
}
\vspace{-4mm}
\end{table*}

\minisection{Overall Component Analysis.}
We progressively add key components to the baseline. \textbf{B} trains directly with initial SAM2 coarse masks, \textbf{Evo} adds Stage 1 bidirectional evolution, \textbf{Ref} adds reliability-guided reference selection, \textbf{Rel} adds reliability-aware loss, and \textbf{RPTM} gives the full ARTEMIS.
As shown in \cref{tab:overall_components}, the gains follow the design logic of ARTEMIS. \circlednum{1} Evo brings the largest improvement, raising scribble Dice from 57.0\%/46.0\%/49.0\%/42.9\% to 77.7\%/63.4\%/71.8\%/63.7\% on Easy-Seen/Easy-Unseen/Hard-Seen/Hard-Unseen. \circlednum{2} Ref further improves Dice to 79.7\%/64.3\%/74.5\%/65.0\%. \circlednum{3} Rel raises it to 81.8\%/64.5\%/75.8\%/65.7\% by down-weighting unreliable supervision. \circlednum{4} RPTM achieves the final Dice of 85.2\%/66.7\%/78.8\%/68.7\%, showing the benefit of transported reference identity.
The 1/8 labeled-data setting shows a similar trend. \circlednum{1} Evo improves Dice from 66.0\%/56.3\%/61.6\%/55.3\% to 80.0\%/61.1\%/72.1\%/63.7\%. \circlednum{2} Ref is most beneficial on Easy-Unseen and Hard-Seen, while \circlednum{3} Rel improves Hard-Unseen Dice from 64.1\% to 67.2\%. \circlednum{4} RPTM further raises Dice from 84.4\%/66.1\%/77.3\%/67.2\% to 86.5\%/68.2\%/80.0\%/70.9\%. Overall, Stage 1 improves pseudo-label quality, and Stage 2 enhances robustness via reference selection, noise suppression, and identity transport.

\minisection{Stage 1: Agent-guided Reliable Anchor Selection.}
We first ablate temporal anchor selection. \textbf{NS} uses every SAM2 coarse mask as an anchor, \textbf{PS} applies SEE-style policy-based selection with area, confidence, and temporal-continuity rules, and \textbf{AS} uses agent judgment before T-NMS. As shown in \cref{tab:s1_selection}, \circlednum{1} NS reaches only 59.8\%/59.9\% Dice on SUN-SEG-Easy-Unseen/Hard-Unseen because unreliable masks are propagated. \circlednum{2} PS raises Dice to 63.4\%/64.7\%, showing that filtering low-quality anchors is necessary. \circlednum{3} AS further improves Dice to 66.7\%/68.7\% and $F_\beta^w$ from 37.5\%/35.1\% to 65.9\%/67.5\% over NS, confirming that semantic agent judgment prevents unreliable anchors in propagation.

\begin{table}[t]
\centering
\caption{Ablation study of reliable anchor selection strategies in Stage 1 on \textbf{SUN-SEG-Easy-Unseen} and \textbf{SUN-SEG-Hard-Unseen} sets. NS: no selection; PS: policy-based selection; AS: agent-guided selection.}
\label{tab:s1_selection}
\vspace{-2mm}
\setlength{\tabcolsep}{5pt} 
\renewcommand{\arraystretch}{0.7}
\resizebox{\linewidth}{!}{ 
\begin{tabular}{l|cccc|cccc}
\toprule
\multirow{2}{*}{Setting} & \multicolumn{4}{c|}{SUN-SEG-Easy-Unseen (\%)} & \multicolumn{4}{c}{SUN-SEG-Hard-Unseen (\%)} \\
& $S_\alpha\uparrow$ & $E_{\phi}^{mn}\uparrow$ & $F_\beta^w\uparrow$ & Dice$\uparrow$ & $S_\alpha\uparrow$ & $E_{\phi}^{mn}\uparrow$ & $F_\beta^w\uparrow$ & Dice$\uparrow$ \\
\midrule
\circlednum{1} NS & 70.4 & 72.1 & 37.5 & 59.8 & 68.8 & 72.3 & 35.1 & 59.9 \\
\circlednum{2} PS & 74.5 & 77.0 & 45.9 & 63.4 & 73.8 & 78.4 & 44.8 & 64.7 \\
\rowcolor{c1!30}
\makecell[l]{\circlednum{3} AS\\~~(\textbf{Ours})} & 78.6 & 79.7 & 65.9 & 66.7 & 79.6 & 81.6 & 67.5 & 68.7 \\
\bottomrule
\end{tabular}
}
\vspace{-6mm}
\end{table}

\minisection{Stage 1: Bidirectional Mask Evolution.}
We then compare evolution directions. \textbf{CM} uses original SAM2 masks without refinement, \textbf{CM+FR} propagates selected anchors forward, and \textbf{CM+BiR} propagates them bidirectionally. In \cref{tab:s1_evolution}, \circlednum{1} CM obtains 54.9\%/53.4\% Dice, showing that isolated sparse-prompt masks lack temporal consistency. \circlednum{2} CM+FR improves Dice to 63.4\%/64.4\%, confirming that reliable anchors can refine later frames. \circlednum{3} CM+BiR reaches 66.7\%/68.7\% Dice, with relative gains of 21.5\%/28.7\% over CM and large $F_\beta^w$ gains, confirming that reverse propagation complements forward propagation and reduces one-directional drift.

\begin{table}[t]
\centering
\caption{Ablation study of mask evolution strategies in Stage 1 on \textbf{SUN-SEG-Easy-Unseen} and \textbf{SUN-SEG-Hard-Unseen} sets. CM: coarse mask; FR: forward refinement; BiR: bidirectional refinement.}
\label{tab:s1_evolution}
\vspace{-2mm}
\setlength{\tabcolsep}{3.5pt} 
\renewcommand{\arraystretch}{0.7}
\resizebox{\linewidth}{!}{ 
\begin{tabular}{l|cccc|cccc}
\toprule
\multirow{2}{*}{Setting} & \multicolumn{4}{c|}{SUN-SEG-Easy-Unseen (\%)} & \multicolumn{4}{c}{SUN-SEG-Hard-Unseen (\%)} \\
& $S_\alpha\uparrow$ & $E_{\phi}^{mn}\uparrow$ & $F_\beta^w\uparrow$ & Dice$\uparrow$ & $S_\alpha\uparrow$ & $E_{\phi}^{mn}\uparrow$ & $F_\beta^w\uparrow$ & Dice$\uparrow$ \\
\midrule
\circlednum{1} CM & 66.5 & 68.3 & 31.9 & 54.9 & 64.3 & 67.2 & 29.1 & 53.4 \\
\circlednum{2} CM+FR & 74.8 & 76.6 & 45.9 & 63.4 & 73.7 & 77.5 & 43.5 & 64.4 \\
\rowcolor{c1!30}
\makecell[l]{\circlednum{3} CM+BiR\\~~(\textbf{Ours})} & 78.6 & 79.7 & 65.9 & 66.7 & 79.6 & 81.6 & 67.5 & 68.7  \\
\bottomrule
\end{tabular}
}
\vspace{-4mm}
\end{table}

\minisection{Stage 1: Pseudo-label Quality.}
We also measure pseudo-label quality after Stage 1. Ground truth is used only for this offline analysis. In \cref{tab:s1_quality}, \circlednum{1} coarse masks obtain 71.0\%/73.4\% Dice and 8.8\%/8.3\% MAE under scribble/1/8 labeled-data settings, showing the noise in initial SAM2 masks. \circlednum{2} Forward evolution raises Dice to 81.5\%/84.2\% and lowers MAE to 4.0\%/3.8\%. \circlednum{3} Bidirectional evolution further improves Dice to 83.1\%/87.6\% and MAE to 3.6\%/1.8\%, explaining the downstream gains from more clear pseudo masks.

\begin{table}[t]
\centering
\caption{Pseudo-label quality analysis after Stage 1 evolution under \textit{Scribble Supervision} and \textit{1/8 Labeled Training Data} settings.}
\label{tab:s1_quality}
\vspace{-2mm}
\setlength{\tabcolsep}{4.5pt}
\renewcommand{\arraystretch}{0.7}
\resizebox{\linewidth}{!}{
\begin{tabular}{l|cccccc}
\toprule
\multirow{2}{*}{Pseudo Label} & \multicolumn{6}{c}{SUN-SEG-Train (\%)} \\
\cmidrule(lr){2-7}
& $S_\alpha\uparrow$ & $E_{\phi}^{mn}\uparrow$ & $F_\beta^w\uparrow$ & Dice$\uparrow$ & IoU$\uparrow$ & MAE$\downarrow$ \\
\midrule

\multicolumn{7}{l}{\textit{Scribble Supervision}} \\
\midrule
\circlednum{1} Coarse mask & 80.6 & 81.3 & 68.9 & 71.0 & 63.0 & 8.8 \\
\circlednum{2} Forward evolved & 87.7 & 91.4 & 80.2 & 81.5 & 73.8 & 4.0 \\
\rowcolor{c1!30}
\makecell{\circlednum{3} Bidirectional evolved\\~(\textbf{Ours})} & 88.6 & 93.0 & 82.0 & 83.1 & 74.5 & 3.6 \\
\midrule

\multicolumn{7}{l}{\textit{1/8 Labeled Training Data}} \\
\midrule
\circlednum{1} Coarse mask & 81.5 & 84.2 & 69.8 & 73.4 & 64.1 & 8.3 \\
\circlednum{2} Forward evolved & 88.8 & 92.5 & 82.1 & 84.2 & 76.1 & 3.8 \\
\rowcolor{c1!30}
\makecell{\circlednum{3} Bidirectional evolved\\~(\textbf{Ours})} & 91.9 & 94.7 & 86.2 & 87.6 & 80.9 & 1.8 \\
\bottomrule
\end{tabular}
}
\vspace{-6mm}
\end{table}

\minisection{Stage 2: Reliability-guided Reference Selection.}
We compare three reference strategies: \textbf{FF} always uses the first frame, \textbf{RR} samples randomly, and \textbf{RF} selects references using the frame-level quality score $Q_t$. As shown in \cref{tab:s2_reference}, \circlednum{1} FF reaches 65.3\%/66.9\% Dice, but the first frame is not always reliable under blur or occlusion. \circlednum{2} RR increases $F_\beta^w$ but lowers Dice, suggesting inconsistent identity cues. \circlednum{3} RF achieves the best Dice of 66.7\%/68.7\% and improves $F_\beta^w$ from 62.2\%/62.5\% to 65.9\%/67.5\% over RR, verifying that $Q_t$ selects a more trustworthy identity source for RPTM.

\begin{table}[t]
\centering
\caption{Ablation study of reference selection strategies in Stage 2 on \textbf{SUN-SEG-Easy-Unseen} and \textbf{SUN-SEG-Hard-Unseen} sets. FF: fixed first frame; RR: random reference; RF: reliability-guided reference frame.}
\label{tab:s2_reference}
\vspace{-2mm}
\setlength{\tabcolsep}{5pt} 
\renewcommand{\arraystretch}{0.7}
\resizebox{\linewidth}{!}{ 
\begin{tabular}{l|cccc|cccc}
\toprule
\multirow{2}{*}{Setting} & \multicolumn{4}{c|}{SUN-SEG-Easy-Unseen (\%)} & \multicolumn{4}{c}{SUN-SEG-Hard-Unseen (\%)} \\
& $S_\alpha\uparrow$ & $E_{\phi}^{mn}\uparrow$ & $F_\beta^w\uparrow$ & Dice$\uparrow$ & $S_\alpha\uparrow$ & $E_{\phi}^{mn}\uparrow$ & $F_\beta^w\uparrow$ & Dice$\uparrow$ \\
\midrule
\circlednum{1} FF & 76.9 & 78.8 & 50.2 & 65.3 & 77.1 & 80.1 & 50.8 & 66.9 \\
\circlednum{2} RR & 76.6 & 76.1 & 62.2 & 62.9 & 77 & 77.6 & 62.5 & 63.3 \\
\rowcolor{c1!30}
\makecell[l]{\circlednum{3} RF\\~~(\textbf{Ours})} & 78.6 & 79.7 & 65.9 & 66.7 & 79.6 & 81.6 & 67.5 & 68.7 \\
\bottomrule
\end{tabular}
}
\vspace{-6mm}
\end{table}

\minisection{Stage 2: Reliability-aware Robust Loss.}
\cref{tab:s2_loss} ablates the robust loss. \textbf{DPS} applies uniform supervision, \textbf{Conf} uses $R_t^{conf}$, \textbf{BiC} adds $R_t^{bidir}$, and \textbf{SJW} adds frame-level judge weights. The results show a progressive reliability effect: \circlednum{1} DPS obtains 63.1\%/64.4\% Dice, revealing the limitation of treating all pseudo labels equally. \circlednum{2} Conf improves Dice to 64.3\%/65.8\%, and \circlednum{3} adding BiC further raises it to 65.8\%/67.4\%. \circlednum{4} Adding SJW gives the best Dice of 66.7\%/68.7\% and improves $F_\beta^w$ from 46.3\%/43.9\% to 65.9\%/67.5\% over DPS, showing that pixel-, temporal-, and frame-level reliability jointly suppress noisy supervision.

\begin{table}[t]
\centering
\caption{Ablation study of reliability-aware robust loss in Stage 2 on \textbf{SUN-SEG-Easy-Unseen} and \textbf{SUN-SEG-Hard-Unseen} sets. DPS: direct pseudo-label supervision; Conf: $R_t^{conf}$; BiC: $R_t^{bidir}$; SJW: frame-level judge weight.}
\label{tab:s2_loss}
\vspace{-2mm}
\setlength{\tabcolsep}{2pt} 
\renewcommand{\arraystretch}{0.7}
\resizebox{\linewidth}{!}{ 
\begin{tabular}{l|cccc|cccc}
\toprule
\multirow{2}{*}{Setting} & \multicolumn{4}{c|}{SUN-SEG-Easy-Unseen (\%)} & \multicolumn{4}{c}{SUN-SEG-Hard-Unseen (\%)} \\
& $S_\alpha\uparrow$ & $E_{\phi}^{mn}\uparrow$ & $F_\beta^w\uparrow$ & Dice$\uparrow$ & $S_\alpha\uparrow$ & $E_{\phi}^{mn}\uparrow$ & $F_\beta^w\uparrow$ & Dice$\uparrow$ \\
\midrule
\circlednum{1} DPS & 74.9 & 77.2 & 46.3 & 63.1 & 74.0 & 78.3 & 43.9 & 64.4 \\
\circlednum{2} Conf & 77.3 & 78.4 & 51.4 & 64.3 & 76.8 & 79.7 & 50.7 & 65.8 \\
\circlednum{3} Conf+BiC & 77.6 & 78.7 & 54.8 & 65.8 & 78.5 & 80.3 & 55.1 & 67.4 \\
\rowcolor{c1!30}
\makecell{\circlednum{4} Conf+BiC+SJW\\(\textbf{Ours})} & 78.6 & 79.7 & 65.9 & 66.7 & 79.6 & 81.6 & 67.5 & 68.7 \\
\bottomrule
\end{tabular}
}
\vspace{-4mm}
\end{table}

\minisection{Stage 2: Reference Prototype Transport Module.}
\cref{tab:rptm_module} ablates RPTM. \circlednum{1} Removing relation-gated injection (\textbf{w/o RG}) drops Dice to 66.4\%/67.7\%, indicating that directly injecting transported identity may introduce noise. \circlednum{2} One-directional evolution (\textbf{Uni-M}) obtains 66.6\%/67.1\% Dice, showing limited robustness to temporal drift. \circlednum{3} The full design (\textbf{Bi-M}) achieves the best Dice of 66.7\%/68.7\% and $F_\beta^w$, validating relation-gated bidirectional identity modeling.

\begin{table}[t]
\centering
\caption{Ablation study of RPTM components on \textbf{SUN-SEG-Easy-Unseen} and \textbf{SUN-SEG-Hard-Unseen} sets. RG: relation gate; Uni-M/Bi-M: unidirectional/bidirectional Mamba.}
\label{tab:rptm_module}
\vspace{-2mm}
\setlength{\tabcolsep}{3.5pt} 
\renewcommand{\arraystretch}{0.7}
\resizebox{\linewidth}{!}{ 
\begin{tabular}{l|cccc|cccc}
\toprule
\multirow{2}{*}{Setting} & \multicolumn{4}{c|}{SUN-SEG-Easy-Unseen (\%)} & \multicolumn{4}{c}{SUN-SEG-Hard-Unseen (\%)} \\
& $S_\alpha\uparrow$ & $E_{\phi}^{mn}\uparrow$ & $F_\beta^w\uparrow$ & Dice$\uparrow$ & $S_\alpha\uparrow$ & $E_{\phi}^{mn}\uparrow$ & $F_\beta^w\uparrow$ & Dice$\uparrow$ \\
\midrule
\circlednum{1} w/o RG & 77.8 & 79.1 & 64.7 & 66.4 & 79.1 & 81.4 & 66.8 & 67.7 \\
\circlednum{2} Uni-M & 78.1 & 79.0 & 65.1 & 66.6 & 78.7 & 81.0 & 66.5 & 67.1 \\
\rowcolor{c1!30}
\makecell[l]{\circlednum{3} Bi-M\\~~(\textbf{Ours})} & 78.6 & 79.7 & 65.9 & 66.7 & 79.6 & 81.6 & 67.5 & 68.7 \\
\bottomrule
\end{tabular}
}
\vspace{-4mm}
\end{table}

\minisection{Stage 2: Reference Identity Token Number.}
\cref{tab:rptm_tokens} analyzes the number of reference identity tokens $N$. \circlednum{1} With one token, RPTM obtains 64.5\%/65.9\% Dice and 46.0\%/44.9\% $F_\beta^w$, showing limited appearance capacity. \circlednum{2} Increasing $N$ to 2 improves $F_\beta^w$, because multiple tokens encode richer identity patterns. \circlednum{3} $N=4$ achieves the best Dice of 66.7\%/68.7\%. \circlednum{4} $N=8$ degrades Dice to 64.3\%/67.0\%, suggesting redundant or noisy cues. Thus, $N=4$ provides the best balance between identity capacity and noise robustness.

\begin{table}[t]
\centering
\caption{Stage 2 parameter analysis of the reference identity token number ($N$) in RPTM on \textbf{SUN-SEG-Easy-Unseen} and \textbf{SUN-SEG-Hard-Unseen} sets.}
\label{tab:rptm_tokens}
\vspace{-2mm}
\setlength{\tabcolsep}{3.5pt} 
\renewcommand{\arraystretch}{0.7}
\resizebox{\linewidth}{!}{ 
\begin{tabular}{l|cccc|cccc}
\toprule
\multirow{2}{*}{$N$} & \multicolumn{4}{c|}{SUN-SEG-Easy-Unseen (\%)} & \multicolumn{4}{c}{SUN-SEG-Hard-Unseen (\%)} \\
& $S_\alpha\uparrow$ & $E_{\phi}^{mn}\uparrow$ & $F_\beta^w\uparrow$ & Dice$\uparrow$ & $S_\alpha\uparrow$ & $E_{\phi}^{mn}\uparrow$ & $F_\beta^w\uparrow$ & Dice$\uparrow$ \\
\midrule
\circlednum{1} 1 & 74.8 & 77.4 & 46.0 & 64.5 & 74.3 & 78.7 & 44.9 & 65.9 \\
\circlednum{2} 2 & 77.7 & 77.7 & 63.7 & 64.5 & 78.9 & 80.2 & 65.0 & 66.1 \\
\rowcolor{c1!30}
\circlednum{3} 4 \textbf{(Ours)} & 78.6 & 79.7 & 65.9 & 66.7 & 79.6 & 81.6 & 67.5 & 68.7 \\
\circlednum{4} 8 & 77.6 & 77.6 & 63.4 & 64.3 & 79.3 & 80.8 & 65.7 & 67.0 \\
\bottomrule
\end{tabular}
}
\vspace{-2mm}
\end{table}

\subsection{Further Analysis}
\begin{figure}
\vspace{-4mm}
    \centering
    \includegraphics[width=\linewidth]{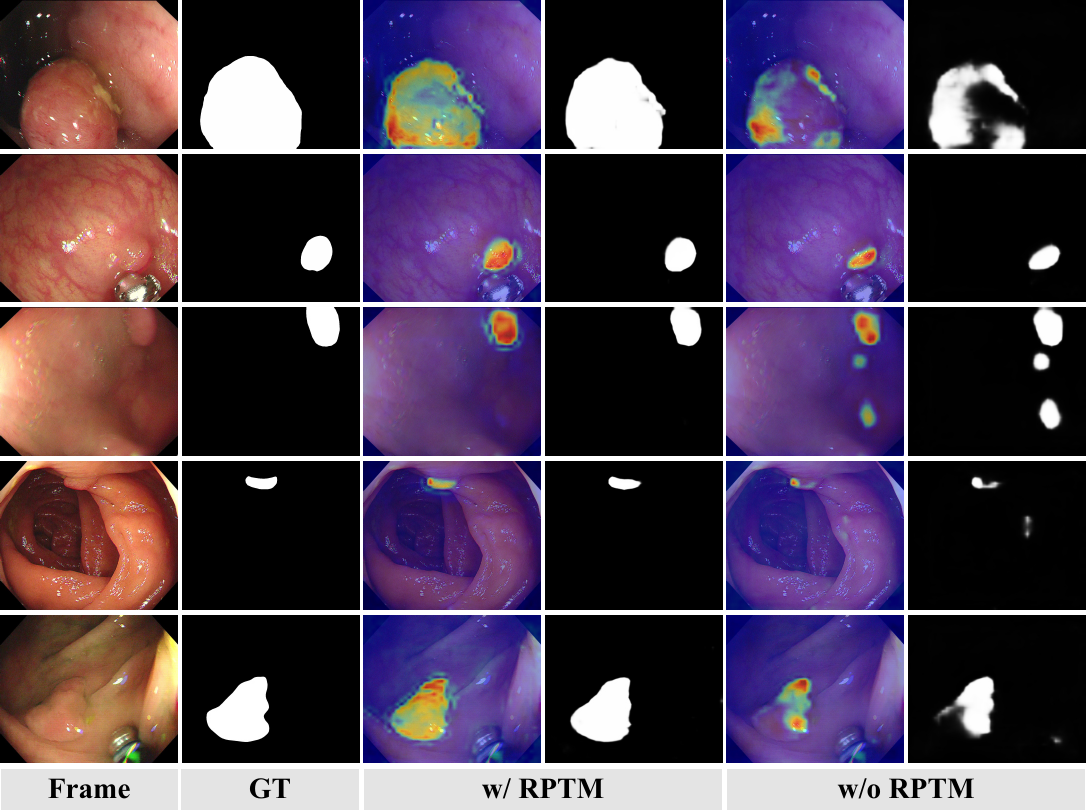}
    \vspace{-6mm}
    \caption{\textbf{Activation visualization of w/o RPTM and RPTM.} Under \textit{Scribble Supervision}, RPTM transports reliable reference-frame identity tokens to the current frame, enhancing polyp-region responses and suppressing distractors.}
    \label{fig:heatmap_weakly_sup}
    \vspace{-4mm}
\end{figure}

\begin{figure}
    \centering
    \includegraphics[width=\linewidth]{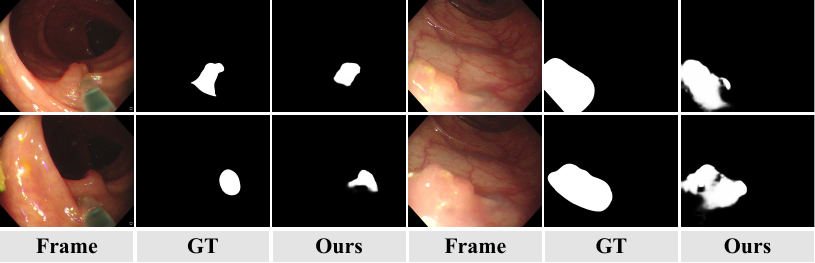}
    \vspace{-6mm}
    \caption{\textbf{Failure cases.} Under \textit{Scribble Supervision}, severe distractors, rapid motion, and uneven illumination may still affect fine boundary details and cause incomplete segmentation.}
    \label{fig:failure_cases}
    \vspace{-4mm}
\end{figure}

\minisection{Activation visualization.}
\cref{fig:heatmap_weakly_sup} visualizes activation maps with and without RPTM under scribble supervision. Without RPTM, the model may activate mucosa-like background or miss ambiguous polyp interiors. With RPTM, reliable reference identity tokens strengthen true-polyp responses and suppress distractors. This indicates that reference prototypes improve semantic discriminability rather than merely smoothing predictions temporally.

\minisection{Failure cases.}
\cref{fig:failure_cases} shows typical failures under scribble supervision. Severe distractors, drastic motion, strong highlights, or uneven illumination may still cause inaccurate boundaries or incomplete masks. These cases motivate more motion-robust temporal modeling and boundary-aware refinement.


\minisection{Future work.}
We will extend ARTEMIS to real-world clinical scenarios and long videos, where complex motion, highlights, low-contrast boundaries, and diverse imaging protocols require stronger temporal reasoning and uncertainty-aware reference selection. We will also explore \textit{memory-efficient reference management}, \textit{drift-resistant propagation}, and \textit{adaptive human-in-the-loop correction} for stable long-video segmentation.

\section{Conclusion}
In this paper, we presented \textbf{ARTEMIS}, a unified framework for imperfectly supervised VPS. Rather than treating weak supervision and semi-supervision as isolated regimes, ARTEMIS completes sparse or missing annotations into temporally consistent dense pseudo masks. Consistent with the weak/semi-supervised setting, ARTEMIS mitigates geometry-degraded pseudo masks, temporally underused pseudo labeling, and reliability-blind pseudo supervision through agent-guided bidirectional mask evolution and reliability-aware robust learning, including reference selection and RPTM. By selecting reliable temporal anchors and propagating them across frames, ARTEMIS better exploits foundation-model priors while suppressing noisy supervision. Experiments on SUN-SEG and CVC-ClinicDB-612 demonstrate state-of-the-art performance under scribble, point, and limited-label supervision. These results show that reliable temporal completion can make low-cost annotations more effective for clinically practical VPS.

\section*{Acknowledgment}
Tong Wang, Yaolei Qi, and Guanyu Yang are supported by the National Natural Science Foundation of China (T2541064, 82441021) and the Jiangsu Province Cutting-edge Technology R\&D Program (BF2025628). This work is also supported by the Big Data Computing Center of Southeast University.

\bibliographystyle{IEEEtran}
\bibliography{egbib}

\newpage
\vskip -12mm

\begin{IEEEbiography}[{\includegraphics[width=1in,height=1.25in,clip,keepaspectratio]{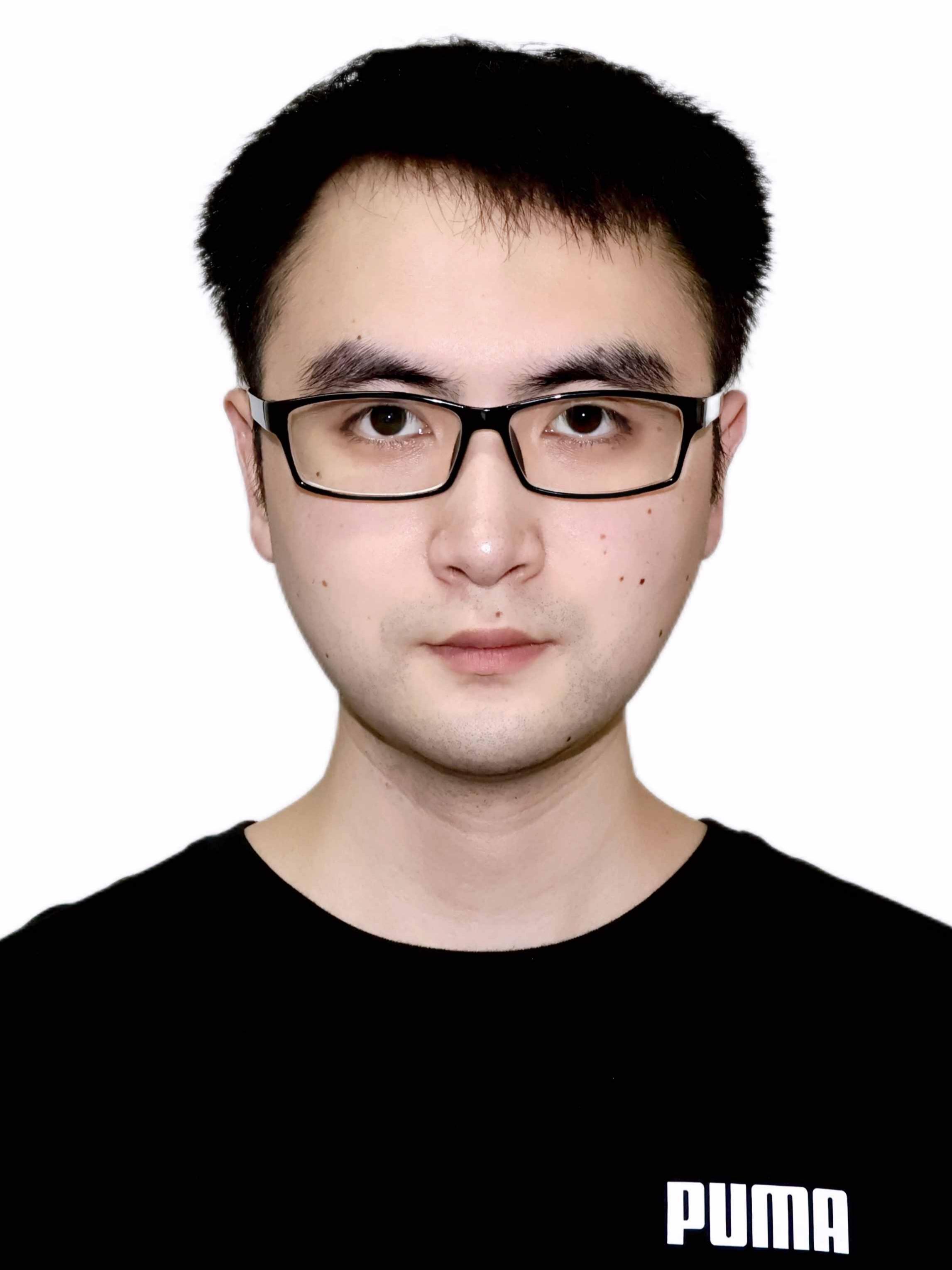}}]{Tong Wang}
is currently a Ph.D. candidate at the Key Laboratory of New Generation Artificial Intelligence Technology and Its Interdisciplinary Applications (Ministry of Education), Southeast University, Nanjing, China. He is also a visiting student at Mohamed bin Zayed University of Artificial Intelligence (MBZUAI). His research interests include medical image analysis and multimodal models.
\end{IEEEbiography}

\vskip -12mm

\begin{IEEEbiography}[{\includegraphics[width=1in,height=1.25in,clip,keepaspectratio]{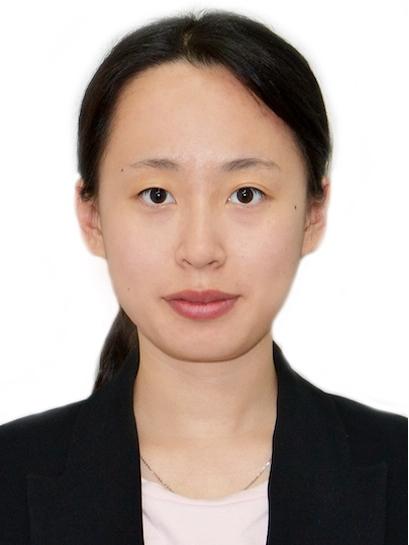}}]{Siwen Wang}
received the B.S. and M.S. degrees in Information and Communication Engineering from Dalian University of Technology, Dalian, China, in 2017 and 2019, respectively. She is currently a Ph.D. candidate at Mohamed bin Zayed University of Artificial Intelligence (MBZUAI). Her research interests include computer vision, pattern recognition, and medical image analysis.
\end{IEEEbiography}

\vskip -12mm

\begin{IEEEbiography}[{\includegraphics[width=1in,height=1.25in,clip,keepaspectratio]{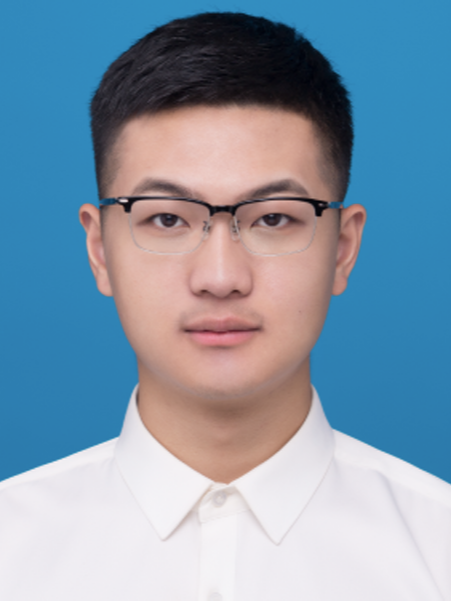}}]{Yaolei Qi}
received the B.E. degree from Southeast University, Nanjing, China, in 2019, where he is currently pursuing the Ph.D. degree with the Laboratory of Image Science and Technology. His research interests lie in deep learning and AI for medical image processing, focusing on knowledge-driven and data-efficient methods for tubular structure analysis. Specifically, his work involves innovative network design, weakly-supervised learning, and computer-assisted diagnosis.
\end{IEEEbiography}

\vskip -12mm

\begin{IEEEbiography}[{\includegraphics[width=1in,height=1.25in,clip,keepaspectratio]{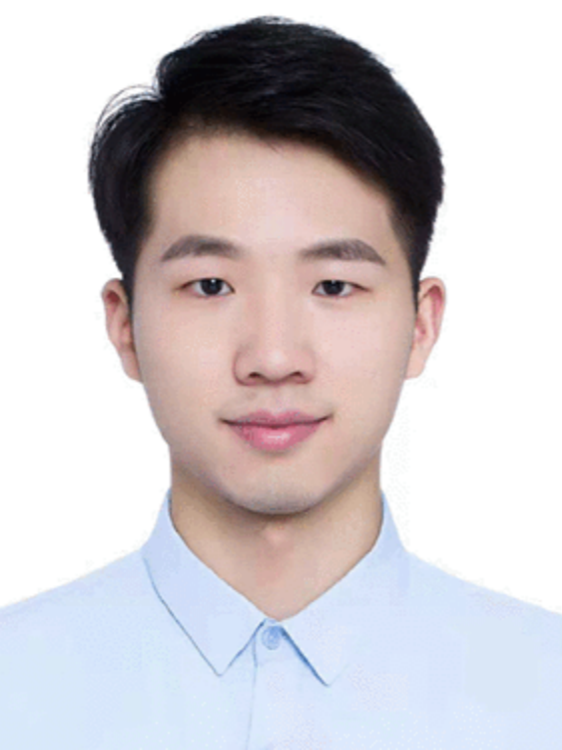}}]{Jinxing Zhou}
received the B.E. and Ph.D. degrees from Hefei University of Technology, China. He is currently a Postdoctoral Researcher with the Department of Computer Vision at Mohamed bin Zayed University of Artificial Intelligence. His research interests include computer vision, audio-visual learning, and multimedia content analysis. He has published over 40 papers in top-tier venues, including IEEE TPAMI, IJCV, IEEE TMM, CVPR, ECCV, ICML, and AAAI.
\end{IEEEbiography}

\vskip -12.5mm

\begin{IEEEbiography}[{\includegraphics[width=1in,height=1.25in,clip,keepaspectratio]{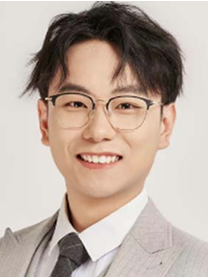}}]{Yuting He}
received B.E. (2018) and Ph.D. (2023) from Hefei University of Technology and Southeast University. Currently a Research Associate at Case Western Reserve University, he visited Western University, Canada (2022). His research focuses on efficient knowledge extraction for computer-aided diagnosis, surgery, and medical imaging. He has published 40+ papers in top venues, including Nature Communications, IEEE TPAMI/TMI, MedIA, CVPR, ICCV, and MICCAI.
\end{IEEEbiography}

\vskip -12.5mm

\begin{IEEEbiography}[{\includegraphics[width=1in,height=1.25in,clip,keepaspectratio]{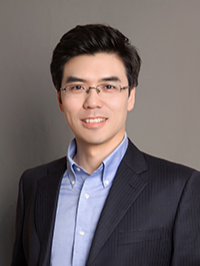}}]{Guanyu Yang}
(Senior Member, IEEE) received B.S. and M.S. degrees in biomedical engineering from Southeast University, China (2002, 2005). He joined the university in 2011 and is currently a Professor and Vice Dean of the School of Computer Science and Engineering. His research focuses on medical image analysis, biomedical processing, and AI for healthcare. He has published 60+ papers in top venues like IEEE TPAMI/TIP/TMI, MedIA, CVPR, ICCV, IJCAI, and MICCAI.
\end{IEEEbiography}

\vskip -12.5mm

\begin{IEEEbiography}[{\includegraphics[width=1in,height=1.25in,clip,keepaspectratio]{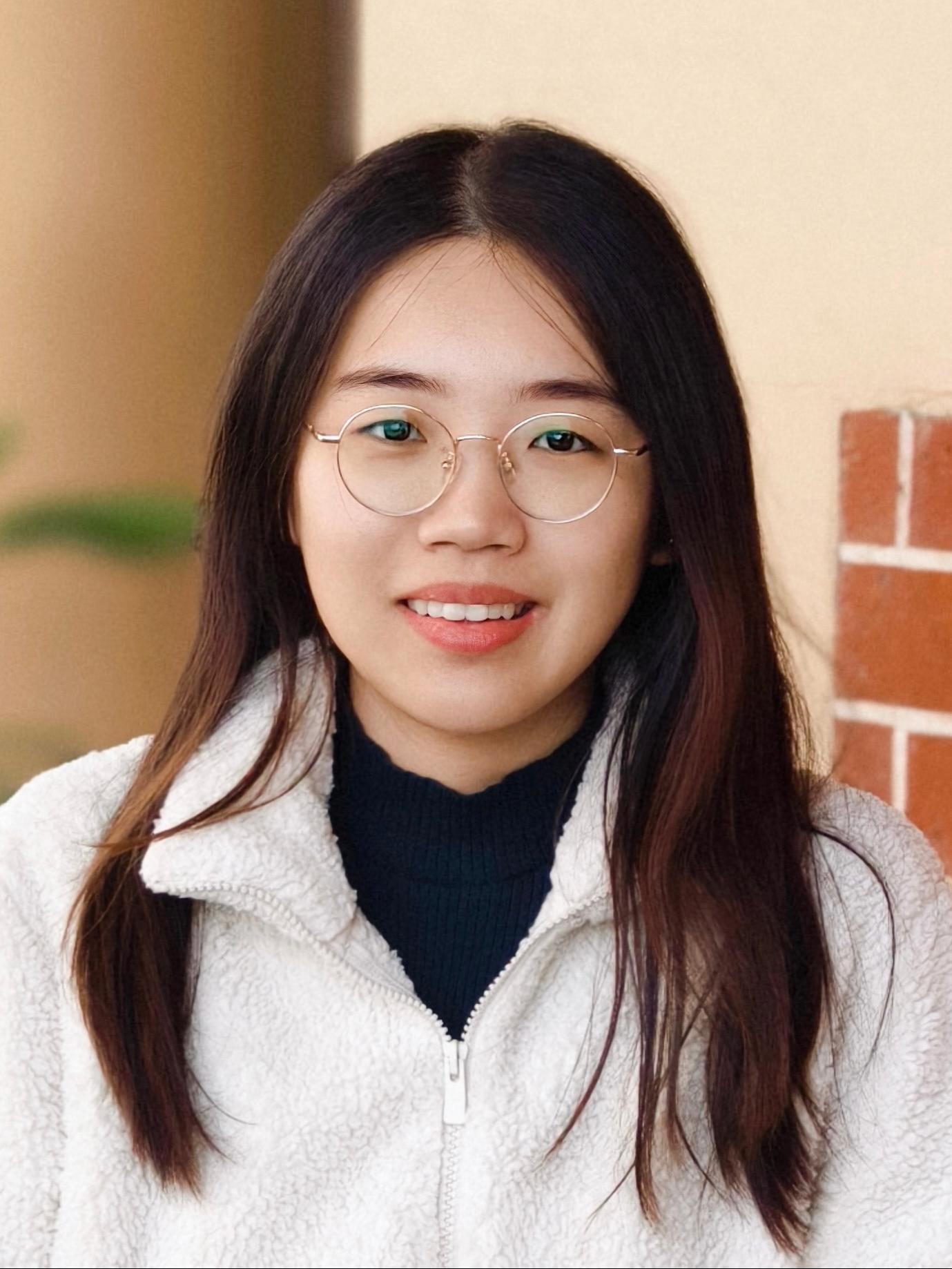}}]{Yutong Xie}
is an Assistant Professor in the Computer Vision Department at MBZUAI. She was previously a Research Fellow at the University of Adelaide’s Australian Institute for Machine Learning. Her research focuses on computer vision and machine learning for healthcare, particularly medical image analysis with limited annotations. She has published 60+ papers in venues including IEEE TPAMI, TMI, TIP, IJCV, MedIA, MICCAI, CVPR, ICCV, ECCV, NeurIPS, ACL, and EMNLP.
\end{IEEEbiography}

\vfill

\end{document}